\documentclass[10pt,twocolumn,letterpaper]{article}

\PassOptionsToPackage{table,xcdraw,dvipsnames}{xcolor}
\usepackage[pagenumbers]{iccv} 
\usepackage[breaklinks,colorlinks,allcolors=iccvblue]{hyperref}

\usepackage{multirow}
\newcommand{\datasetsize}{114k}

\usepackage{pgf}
\usepackage{xcolor}

\definecolor{goodred}{HTML}{e7638d}     
\definecolor{goodyellow}{HTML}{f2e17c}  
\definecolor{goodgreen}{HTML}{a1d76a}   

\newcommand{\definecolorrange}[5]{%
    \expandafter\newcommand\csname colorrange#1#2#3\endcsname[1]{%
        \pgfmathsetmacro{\minval}{#4}  
        \pgfmathsetmacro{\maxval}{#5}  
        \pgfmathsetmacro{\midval}{(\minval+\maxval)/2}  

        \pgfmathsetmacro{\PercentColor}{100.0*(##1-\minval)/(\maxval-\minval)}%
        \ifdim ##1 pt < \minval pt
            \cellcolor{goodred}{##1}%
        \else
            \ifdim ##1 pt > \maxval pt
                \cellcolor{goodgreen}{##1}%
            \else
                \ifdim ##1 pt < \midval pt
                    \pgfmathsetmacro{\PercentLow}{200.0*(##1-\minval)/(\maxval-\minval)}%
                    \xdef\PercentLow{\PercentLow}%
                    \cellcolor{goodyellow!\PercentLow!goodred}{##1}%
                \else
                    \pgfmathsetmacro{\PercentHigh}{200.0*(##1-\midval)/(\maxval-\minval)}%
                    \xdef\PercentHigh{\PercentHigh}%
                    \cellcolor{goodgreen!\PercentHigh!goodyellow}{##1}%
                \fi
            \fi
        \fi
    }
}

\definecolorrange{ATL}{cnn}{bench}{56.07}{58.76}
\definecolorrange{SBM}{cnn}{bench}{53.43}{64.09}
\definecolorrange{ISL}{cnn}{bench}{77.45}{79.14}
\definecolorrange{HNT}{cnn}{bench}{59.37}{66.58}
\definecolorrange{HAN}{cnn}{bench}{45}{55.14}
\definecolorrange{MSF}{cnn}{bench}{52.19}{57.42}
\definecolorrange{TPC}{cnn}{bench}{74.69}{79.94}
\definecolorrange{YBM}{cnn}{bench}{54.42}{60.07}
\definecolorrange{COS}{cnn}{bench}{46.10}{71.57}
\definecolorrange{ACD}{cnn}{bench}{90.72}{92.09}
\definecolorrange{AMO}{cnn}{bench}{83.88}{88.0}
\definecolorrange{KIT}{cnn}{bench}{77.61}{87.48}
\definecolorrange{IDMean}{cnn}{bench}{58.33}{65.11}
\definecolorrange{OODMean}{cnn}{bench}{84.07}{88.30}
\definecolorrange{Mean}{cnn}{bench}{64.77}{70.91}

\definecolorrange{ATL}{trans}{bench}{46.80}{61.16}
\definecolorrange{SBM}{trans}{bench}{34.15}{56.67}
\definecolorrange{ISL}{trans}{bench}{73.29}{77.54}
\definecolorrange{HNT}{trans}{bench}{50.15}{66.12}
\definecolorrange{HAN}{trans}{bench}{46.49}{57.24}
\definecolorrange{MSF}{trans}{bench}{53.27}{56.02}
\definecolorrange{TPC}{trans}{bench}{65.52}{78.31}
\definecolorrange{YBM}{trans}{bench}{43}{54.35}
\definecolorrange{COS}{trans}{bench}{51.16}{72.02}
\definecolorrange{ACD}{trans}{bench}{87.27}{92.16}
\definecolorrange{AMO}{trans}{bench}{65.81}{87.24}
\definecolorrange{KIT}{trans}{bench}{70.21}{86.74}
\definecolorrange{IDMean}{trans}{bench}{52.00}{64.34}
\definecolorrange{OODMean}{trans}{bench}{74.43}{88.69}
\definecolorrange{Mean}{trans}{bench}{57.61}{70.42}

\definecolorrange{ATL}{cnn}{onek}{56.64}{61.04}
\definecolorrange{SBM}{cnn}{onek}{53.43}{62.35}
\definecolorrange{ISL}{cnn}{onek}{77.24}{79.14}
\definecolorrange{HNT}{cnn}{onek}{60.83}{65.75}
\definecolorrange{HAN}{cnn}{onek}{53.37}{59.48}
\definecolorrange{MSF}{cnn}{onek}{52.19}{55.08}
\definecolorrange{TPC}{cnn}{onek}{76.17}{79.94}
\definecolorrange{YBM}{cnn}{onek}{56.12}{60.24}
\definecolorrange{COS}{cnn}{onek}{46.19}{72.24}
\definecolorrange{ACD}{cnn}{onek}{91.10}{92.11}
\definecolorrange{AMO}{cnn}{onek}{88.00}{88.86}
\definecolorrange{KIT}{cnn}{onek}{87.21}{88.24}
\definecolorrange{IDMean}{cnn}{onek}{61.08}{65.6134}
\definecolorrange{OODMean}{cnn}{onek}{88.77}{89.67}
\definecolorrange{Mean}{cnn}{onek}{68.00}{71.63}

\newcommand\offset{1}

\definecolorrange{SBM}{VoCo}{cnn}{70.26-\offset}{73.86}
\definecolorrange{ATL}{VoCo}{cnn}{59.32-\offset}{62.78}
\definecolorrange{AMO}{VoCo}{cnn}{85.91-\offset}{86.98}
\definecolorrange{KIT}{VoCo}{cnn}{84.08-\offset}{85.80}
\definecolorrange{AVG}{VoCo}{cnn}{74.92-\offset}{77.09}
\definecolorrange{SBM}{SwinUNETR}{cnn}{69.16-\offset}{71.05}
\definecolorrange{ATL}{SwinUNETR}{cnn}{60.23-\offset}{61.17}
\definecolorrange{AMO}{SwinUNETR}{cnn}{85.42-\offset}{86.61}
\definecolorrange{KIT}{SwinUNETR}{cnn}{81.77-\offset}{83.83}
\definecolorrange{AVG}{SwinUNETR}{cnn}{74.44-\offset}{75.31}
\definecolorrange{SBM}{SimCLR}{cnn}{72.79-\offset}{73.94}
\definecolorrange{ATL}{SimCLR}{cnn}{57.95-\offset}{62.36}
\definecolorrange{AMO}{SimCLR}{cnn}{86.72-\offset}{87.12}
\definecolorrange{KIT}{SimCLR}{cnn}{81.20-\offset}{83.10}
\definecolorrange{AVG}{SimCLR}{cnn}{74.48-\offset}{76.30}
\definecolorrange{SBM}{VF}{cnn}{72.57-\offset}{75.81}
\definecolorrange{ATL}{VF}{cnn}{63.20-\offset}{65.07}
\definecolorrange{AMO}{VF}{cnn}{86.05-\offset}{86.58}
\definecolorrange{KIT}{VF}{cnn}{83.51-\offset}{85.41}
\definecolorrange{AVG}{VF}{cnn}{76.60-\offset}{77.84}
\definecolorrange{SBM}{MG}{cnn}{73.37-\offset}{74.68}
\definecolorrange{ATL}{MG}{cnn}{61.23-\offset}{63.46}
\definecolorrange{AMO}{MG}{cnn}{86.93-\offset}{87.34}
\definecolorrange{KIT}{MG}{cnn}{83.48-\offset}{84.15}
\definecolorrange{AVG}{MG}{cnn}{76.62-\offset}{77.36}
\definecolorrange{SBM}{MAE}{cnn}{71.84-\offset}{74.79}
\definecolorrange{ATL}{MAE}{cnn}{62.78-\offset}{63.61}
\definecolorrange{AMO}{MAE}{cnn}{86.78-\offset}{87.62}
\definecolorrange{KIT}{MAE}{cnn}{83.48-\offset}{85.24}
\definecolorrange{AVG}{MAE}{cnn}{76.49-\offset}{77.75}
\definecolorrange{SBM}{SD}{cnn}{72.40-\offset}{75.47}
\definecolorrange{ATL}{SD}{cnn}{62.16-\offset}{63.76}
\definecolorrange{AMO}{SD}{cnn}{86.45-\offset}{87.54}
\definecolorrange{KIT}{SD}{cnn}{82.20-\offset}{83.78}
\definecolorrange{AVG}{SD}{cnn}{76.40-\offset}{76.87}
\definecolorrange{SBM}{Mean}{cnn}{72.82-\offset}{73.23}
\definecolorrange{ATL}{Mean}{cnn}{61.68-\offset}{62.40}
\definecolorrange{AMO}{Mean}{cnn}{86.42-\offset}{86.69}
\definecolorrange{KIT}{Mean}{cnn}{83.32-\offset}{84.09}
\definecolorrange{AVG}{Mean}{cnn}{76.15-\offset}{76.60}

\definecolorrange{SBM}{VoCo}{TR}{37.55-\offset}{44.16}
\definecolorrange{ATL}{VoCo}{TR}{48.04-\offset}{48.53}
\definecolorrange{AMO}{VoCo}{TR}{66.01-\offset}{67.71}
\definecolorrange{KIT}{VoCo}{TR}{66.33-\offset}{68.31}
\definecolorrange{AVG}{VoCo}{TR}{54.64-\offset}{57.98}
\definecolorrange{SBM}{SwinUNETR}{TR}{41.64-\offset}{47.65}
\definecolorrange{ATL}{SwinUNETR}{TR}{47.72-\offset}{48.87}
\definecolorrange{AMO}{SwinUNETR}{TR}{66.51-\offset}{67.96}
\definecolorrange{KIT}{SwinUNETR}{TR}{68.00-\offset}{69.13}
\definecolorrange{AVG}{SwinUNETR}{TR}{56.42-\offset}{58.10}
\definecolorrange{SBM}{SimCLR}{TR}{54.11-\offset}{59.92}
\definecolorrange{ATL}{SimCLR}{TR}{51.97-\offset}{56.60}
\definecolorrange{AMO}{SimCLR}{TR}{75.88-\offset}{76.42}
\definecolorrange{KIT}{SimCLR}{TR}{77.47-\offset}{80.58}
\definecolorrange{AVG}{SimCLR}{TR}{65.38-\offset}{68.38}
\definecolorrange{SBM}{VF}{TR}{64.12-\offset}{66.03}
\definecolorrange{ATL}{VF}{TR}{63.00-\offset}{63.66}
\definecolorrange{AMO}{VF}{TR}{84.81-\offset}{85.73}
\definecolorrange{KIT}{VF}{TR}{80.40-\offset}{83.02}
\definecolorrange{AVG}{VF}{TR}{73.14-\offset}{74.31}
\definecolorrange{SBM}{MG}{TR}{59.11-\offset}{62.95}
\definecolorrange{ATL}{MG}{TR}{57.81-\offset}{59.43}
\definecolorrange{AMO}{MG}{TR}{82.91-\offset}{83.12}
\definecolorrange{KIT}{MG}{TR}{80.00-\offset}{82.00}
\definecolorrange{AVG}{MG}{TR}{70.14-\offset}{71.51}
\definecolorrange{SBM}{MAE}{TR}{66.88-\offset}{70.86}
\definecolorrange{ATL}{MAE}{TR}{60.95-\offset}{61.48}
\definecolorrange{AMO}{MAE}{TR}{87.80-\offset}{87.93}
\definecolorrange{KIT}{MAE}{TR}{86.29-\offset}{86.45}
\definecolorrange{AVG}{MAE}{TR}{75.39-\offset}{76.78}
\definecolorrange{SBM}{SimMIM}{TR}{66.16-\offset}{68.10}
\definecolorrange{ATL}{SimMIM}{TR}{59.80-\offset}{60.74}
\definecolorrange{AMO}{SimMIM}{TR}{87.79-\offset}{88.22}
\definecolorrange{KIT}{SimMIM}{TR}{83.78-\offset}{85.54}
\definecolorrange{AVG}{SimMIM}{TR}{74.50-\offset}{75.58}
\definecolorrange{SBM}{Mean}{TR}{56.30-\offset}{59.67}
\definecolorrange{ATL}{Mean}{TR}{56.07-\offset}{56.76}
\definecolorrange{AMO}{Mean}{TR}{78.90-\offset}{79.51}
\definecolorrange{KIT}{Mean}{TR}{77.86-\offset}{79.08}
\definecolorrange{AVG}{Mean}{TR}{67.42-\offset}{68.76}

\definecolorrange{ATL}{cnn}{data}{58.25}{59.33}
\definecolorrange{SBM}{cnn}{data}{62.41}{65.21}
\definecolorrange{ISL}{cnn}{data}{77.89}{78.57}
\definecolorrange{HNT}{cnn}{data}{65.80}{66.58}
\definecolorrange{HAN}{cnn}{data}{51.06}{55.14}
\definecolorrange{MSF}{cnn}{data}{56.24}{57.42}
\definecolorrange{TPC}{cnn}{data}{77.17}{79.00}
\definecolorrange{YBM}{cnn}{data}{59.12}{60.28}
\definecolorrange{COS}{cnn}{data}{68.90}{73.37}
\definecolorrange{ACD}{cnn}{data}{91.98}{92.11}
\definecolorrange{AMO}{cnn}{data}{86.57}{86.78}
\definecolorrange{KIT}{cnn}{data}{85.66}{86.20}
\definecolorrange{IDMean}{cnn}{data}{64.58}{65.31}
\definecolorrange{OODMean}{cnn}{data}{88.13}{88.30}
\definecolorrange{Mean}{cnn}{data}{70.48}{71.05}

\definecolorrange{MRNet}{cnn}{AP}{66.33}{71.99}
\definecolorrange{RSNASpine}{cnn}{AP}{65.39}{71.21}
\definecolorrange{ABIDE}{cnn}{AP}{57.18}{62.32}
\definecolorrange{Mean}{cnn}{AP}{62.97}{68.18}
\definecolorrange{MRNet}{cnn}{BalAcc}{76.61}{80.35}
\definecolorrange{RSNASpine}{cnn}{BalAcc}{62.25}{67.96}
\definecolorrange{ABIDE}{cnn}{BalAcc}{53.29}{60.18}
\definecolorrange{Mean}{cnn}{BalAcc}{65.80}{66.58}
\definecolorrange{MRNet}{tr}{AP}{60.58}{68.86}
\definecolorrange{RSNASpine}{tr}{AP}{58.91}{62.09}
\definecolorrange{ABIDE}{tr}{AP}{52.41}{56.84}
\definecolorrange{Mean}{tr}{AP}{57.45}{62.14}
\definecolorrange{MRNet}{tr}{BalAcc}{75.49}{77.73}
\definecolorrange{RSNASpine}{tr}{BalAcc}{57.10}{61.64}
\definecolorrange{ABIDE}{tr}{BalAcc}{51.28}{56.06}
\definecolorrange{Mean}{tr}{BalAcc}{61.55}{63.94}

\newcommand{\alldatasetcite}[0]{\nocite{ds000001:1.0.0,ds000109:2.0.2,ds000170:00001,ds000220:1.0.0,ds000245:00001,ds001226:5.0.0,ds001365:1.0.0,ds001511:1.0.5,ds001614:1.1.0,ds001780:1.0.0,ds001928:1.1.0,ds002144:2.0.0,ds002295:1.0.0,ds002425:1.0.2,ds002672:1.0.0,ds002738:1.0.2,ds002842:1.0.1,ds003017:1.0.3,ds003126:1.3.1,ds003357:1.0.0,ds003441:1.0.0,ds003500:1.2.0,ds003606:1.0.0,ds003715:1.0.0,ds003799:2.0.0,ds003950:1.0.0,ds004073:1.0.1,ds004146:1.0.4,ds004271:1.1.0,ds004400:1.0.6,ds004489:1.0.1,ds004589:1.0.0,ds004663:1.0.4,ds004746:1.0.1,ds004869:1.1.1,ds004993:1.1.2,ds005088:1.0.0,ds005237:1.1.3,ds005374:1.0.1,ds005525:1.0.0,ds000002:00002,ds000171:00001,ds000246:1.0.1,ds001371:1.1.1,ds001621:1.1.0,ds001942:1.2.0,ds002306:1.1.0,ds002674:1.0.6,ds002843:1.0.1,ds003136:1.0.0,ds003442:1.0.0,ds003612:1.0.4,ds003814:1.0.0,ds004078:1.2.1,ds004274:1.0.0,ds004493:1.0.2,ds004666:1.0.5,ds004873:2.0.6,ds005096:1.0.3,ds005375:1.0.0,ds000003:1.0.0,ds000005:00001,ds000006:00001,ds000007:00001,ds000008:00001,ds000009:00002,ds000011:1.0.0,ds000017:00001,ds000030:1.0.0,ds000031:2.0.2,ds000051:00001,ds000052:00001,ds000053:00001,ds000101:00004,ds000102:00001,ds000105:3.0.0,ds000107:00001,ds000108:00002,ds000110:00001,ds000113:1.3.0,ds000114:1.0.2,ds000115:00001,ds000116:00003,ds000117:1.1.0,ds000119:00001,ds000120:00001,ds000121:00001,ds000122:00001,ds000133:00001,ds000140:00001,ds000144:00002,ds000148:1.0.0,ds000149:1.0.0,ds000157:00001,ds000158:1.0.0,ds000164:00001,ds000168:1.0.0,ds000172:1.0.1,ds000174:1.0.1,ds000200:00001,ds000201:1.0.3,ds000202:00001,ds000203:00001,ds000204:00002,ds000205:00001,ds000206:1.0.0,ds000208:1.0.1,ds000210:00002,ds000212:1.0.0,ds000213:00002,ds000214:00001,ds000216:00001,ds000217:1.0.0,ds000218:00002,ds000219:00001,ds000221:1.0.0,ds000222:1.0.1,ds000223:00001,ds000224:1.0.4,ds000228:1.1.1,ds000229:00001,ds000231:1.0.0,ds000232:00001,ds000233:1.0.1,ds000234:00002,ds000235:2.0.1,ds000236:2.0.1,ds000237:1.0.0,ds000238:00002,ds000239:00001,ds000240:2.0.0,ds000241:00002,ds000243:00001,ds000244:1.0.0,ds000247:1.0.2,ds000248:1.2.4,ds000249:00002,ds000253:00002,ds000254:1.0.0,ds000255:00002,ds000256:00002,ds000258:1.0.1,ds001021:1.0.0,ds001037:00001,ds001105:00001,ds001110:00003,ds001131:1.0.0,ds001132:1.0.0,ds001145:1.0.0,ds001168:1.0.1,ds001178:00002,ds001218:00001,ds001228:00002,ds001229:1.0.1,ds001232:1.0.0,ds001233:00003,ds001235:00001,ds001241:1.0.1,ds001242:1.0.0,ds001246:1.2.1,ds001247:00001,ds001297:00001,ds001299:1.0.0,ds001302:1.0.1,ds001306:1.0.0,ds001338:1.0.0,ds001339:00003,ds001344:1.0.0,ds001345:1.0.0,ds001353:1.0.0,ds001357:00001,ds001379:1.0.0,ds001386:00001,ds001399:2.0.0,ds001408:1.0.3,ds001415:1.0.0,ds001417:1.0.0,ds001419:1.0.1,ds001420:1.2.0,ds001421:1.4.1,ds001430:1.0.2,ds001439:1.2.0,ds001454:1.3.1,ds001486:1.3.1,ds001491:1.0.0,ds001497:1.0.2,ds001499:1.3.1,ds001506:1.3.1,ds001510:2.0.3,ds001517:1.0.3,ds001521:1.0.2,ds001525:1.1.1,ds001534:1.1.0,ds001545:1.1.1,ds001551:1.0.0,ds001553:1.0.1,ds001554:1.0.0,ds001555:1.0.1,ds001563:1.0.1,ds001566:1.0.1,ds001576:1.0.0,ds001590:1.0.1,ds001595:1.0.0,ds001597:1.0.0,ds001600:1.0.0,ds001607:1.0.1,ds001608:1.0.1,ds001612:1.0.2,ds001634:1.0.1,ds001635:1.0.1,ds001652:2.0.0,ds001705:1.0.1,ds001715:1.0.0,ds001722:1.1.0,ds001728:1.1.0,ds001734:1.0.5,ds001740:2.2.0,ds001743:1.0.1,ds001745:1.1.0,ds001747:1.1.0,ds001748:1.0.4,ds001751:1.0.2,ds001761:2.0.1,ds001762:1.0.1,ds001771:1.0.2,ds001775:1.0.1,ds001784:1.1.2,ds001796:1.7.0,ds001814:1.0.7,ds001818:1.0.0,ds001832:1.0.1,ds001838:1.0.1,ds001839:1.0.1,ds001840:1.0.2,ds001847:1.0.1,ds001848:1.0.1,ds001882:1.0.7,ds001883:1.0.3,ds001894:1.4.2,ds001907:3.1.0,ds001912:1.0.2,ds001921:1.0.0,ds001923:1.0.0,ds001926:1.0.1,ds001927:2.1.0,ds001946:1.0.3,ds001972:1.0.2,ds001978:1.0.3,ds001984:1.0.2,ds002000:1.0.0,ds002001:1.0.0,ds002006:1.0.1,ds002011:1.0.0,ds002013:1.0.3,ds002014:1.0.1,ds002016:1.0.0,ds002033:1.0.1,ds002040:1.0.2,ds002041:2.0.0,ds002076:1.0.1,ds002080:4.0.0,ds002105:1.1.0,ds002116:1.0.0,ds002149:1.0.5,ds002153:1.0.1,ds002155:1.0.0,ds002156:2.0.0,ds002158:1.0.2,ds002168:1.1.0,ds002169:1.0.0,ds002185:1.1.0,ds002207:1.0.0,ds002232:1.0.0,ds002236:1.1.1,ds002237:1.1.0,ds002241:1.1.1,ds002242:1.0.0,ds002250:1.0.1,ds002270:1.0.0,ds002278:2.0.0,ds002293:1.0.0,ds002294:1.0.1,ds002311:1.1.0,ds002316:1.0.0,ds002320:1.1.0,ds002322:1.0.4,ds002328:1.0.2,ds002330:1.1.0,ds002336:2.0.2,ds002338:2.0.2,ds002345:1.1.4,ds002363:1.1.1,ds002366:1.1.0,ds002367:1.0.0,ds002380:1.0.1,ds002382:1.0.1,ds002411:1.1.0,ds002419:1.0.3,ds002422:1.1.0,ds002424:1.2.0,ds002513:1.0.0,ds002522:1.0.2,ds002543:1.0.1,ds002547:1.1.0,ds002549:1.0.1,ds002550:1.0.1,ds002574:1.0.1,ds002578:1.1.0,ds002596:1.0.1,ds002603:1.0.0,ds002606:1.1.0,ds002608:1.0.2,ds002609:1.0.3,ds002614:1.0.0,ds002620:1.0.0,ds002634:3.0.0,ds002643:1.1.0,ds002647:1.0.1,ds002655:1.0.1,ds002675:1.0.0,ds002684:1.0.0,ds002685:2.0.0,ds002687:1.2.0,ds002702:1.0.1,ds002711:1.1.0,ds002712:1.0.1,ds002715:1.0.0,ds002717:1.0.1,ds002725:1.0.0,ds002726:1.0.1,ds002727:1.0.2,ds002731:1.0.2,ds002732:1.0.0,ds002733:1.0.1,ds002734:1.0.2,ds002735:1.0.2,ds002737:1.0.1,ds002739:1.0.0,ds002741:1.0.2,ds002743:1.0.1,ds002748:1.0.5,ds002750:1.0.1,ds002766:3.0.2,ds002770:2.0.0,ds002773:1.0.0,ds002776:1.0.2,ds002785:2.0.0,ds002790:2.0.0,ds002793:1.0.1,ds002797:1.0.2,ds002799:1.0.4,ds002813:1.0.0,ds002814:1.3.0,ds002835:1.0.1,ds002837:2.0.0,ds002841:1.0.1,ds002848:1.0.1,ds002872:1.3.0,ds002878:2.0.0,ds002879:1.1.1,ds002886:1.1.0,ds002896:1.0.0,ds002898:1.4.2,ds002905:1.0.1,ds002936:1.0.0,ds002938:1.0.1,ds002940:1.0.1,ds002979:1.0.0,ds002989:1.0.0,ds002994:1.0.3,ds002995:1.0.1,ds003007:1.0.1,ds003011:1.2.3,ds003012:1.0.4,ds003020:3.0.0,ds003037:2.1.0,ds003043:1.0.0,ds003047:1.0.0,ds003059:1.0.0,ds003076:1.0.1,ds003078:1.0.0,ds003082:1.0.1,ds003083:1.0.1,ds003085:1.0.0,ds003089:1.0.1,ds003094:1.0.0,ds003095:1.0.0,ds003096:1.0.2,ds003097:1.2.1,ds003098:1.0.0,ds003103:1.0.1,ds003104:1.0.0,ds003114:1.0.1,ds003138:1.0.1,ds003145:1.0.2,ds003146:1.0.2,ds003148:1.0.1,ds003151:2.0.1,ds003170:2.0.0,ds003171:2.0.1,ds003176:2.0.1,ds003192:1.0.1,ds003216:3.0.11,ds003233:1.2.1,ds003242:1.0.0,ds003338:1.1.0,ds003340:1.0.4,ds003342:1.0.0,ds003345:1.0.2,ds003346:1.1.2,ds003354:1.0.1,ds003358:1.0.0,ds003367:1.0.0,ds003381:1.0.1,ds003382:1.5.0,ds003392:1.0.4,ds003397:1.2.3,ds003401:1.0.1,ds003404:1.0.5,ds003416:2.0.2,ds003424:1.0.0,ds003425:1.0.2,ds003430:1.2.0,ds003433:1.0.1,ds003434:1.0.1,ds003436:1.0.0,ds003437:1.0.2,ds003438:1.0.0,ds003439:1.0.0,ds003440:1.0.0,ds003443:1.0.0,ds003444:1.0.0,ds003445:1.0.0,ds003446:1.0.0,ds003452:1.0.1,ds003453:1.0.4,ds003454:1.0.1,ds003455:1.0.2,ds003459:1.0.2,ds003465:1.0.7,ds003466:1.1.1,ds003469:1.0.0,ds003470:2.0.0,ds003481:1.0.3,ds003484:1.0.0,ds003487:2.0.0,ds003495:1.0.0,ds003499:1.0.1,ds003505:1.1.2,ds003507:1.0.1,ds003508:1.0.0,ds003511:1.1.2,ds003521:2.2.0,ds003524:1.0.0,ds003540:1.0.1,ds003542:1.0.0,ds003545:1.0.0,ds003548:1.0.1,ds003550:1.0.2,ds003553:1.0.2,ds003554:1.0.3,ds003563:1.1.0,ds003568:1.0.4,ds003569:1.0.0,ds003574:1.0.2,ds003592:1.0.13,ds003604:1.0.7,ds003633:1.0.4,ds003639:1.0.0,ds003642:1.1.0,ds003643:2.0.7,ds003653:1.0.0,ds003659:2.0.4,ds003661:1.0.0,ds003669:1.0.0,ds003673:2.0.1,ds003684:1.0.0,ds003688:1.0.7,ds003691:1.0.0,ds003696:1.0.0,ds003701:1.0.1,ds003707:1.0.0,ds003709:1.0.0,ds003711:1.0.0,ds003714:1.0.1,ds003716:1.0.0,ds003717:1.1.0,ds003720:1.0.1,ds003721:1.0.1,ds003745:2.1.1,ds003752:1.0.0,ds003758:1.0.2,ds003763:1.0.5,ds003764:1.0.5,ds003768:1.0.11,ds003770:1.2.1,ds003772:1.0.1,ds003777:1.0.1,ds003778:1.0.0,ds003782:1.0.1,ds003787:1.0.1,ds003789:2.0.0,ds003791:1.0.0,ds003798:1.0.5,ds003823:1.3.5,ds003826:3.0.1,ds003831:1.0.0,ds003834:1.0.2,ds003835:1.0.2,ds003836:1.0.0,ds003848:1.0.3,ds003849:1.0.0,ds003851:1.0.1,ds003858:1.0.1,ds003871:1.0.2,ds003872:1.0.0,ds003877:1.1.1,ds003892:5.0.0,ds003900:1.1.1,ds003922:1.0.1,ds003927:1.0.1,ds003949:1.0.1,ds003965:1.0.0,ds003967:1.0.0,ds003972:1.0.0,ds003974:3.0.0,ds003988:1.0.0,ds003990:1.0.2,ds003993:1.0.0,ds003999:1.0.2,ds004006:1.0.4,ds004007:1.0.2,ds004009:1.0.0,ds004021:1.0.0,ds004024:1.0.1,ds004038:1.0.0,ds004042:1.0.1,ds004044:2.0.3,ds004054:1.0.0,ds004056:1.0.2,ds004065:1.0.0,ds004081:1.0.0,ds004086:1.2.0,ds004091:2.0.0,ds004094:1.0.1,ds004097:1.1.0,ds004101:1.0.1,ds004102:1.0.1,ds004103:1.0.0,ds004107:1.0.0,ds004109:1.0.0,ds004128:1.0.0,ds004129:1.0.0,ds004130:1.0.0,ds004131:1.0.1,ds004134:1.0.1,ds004141:1.0.5,ds004142:1.0.1,ds004144:1.0.2,ds004158:2.0.5,ds004169:1.0.7,ds004173:1.0.2,ds004182:1.0.1,ds004187:1.0.2,ds004192:1.0.7,ds004194:2.0.0,ds004196:2.0.2,ds004199:1.0.6,ds004212:2.0.1,ds004213:1.0.1,ds005752:2.1.0,ds004217:1.0.0,ds004219:1.0.0,ds004226:1.0.0,ds004228:1.0.1,ds004230:3.0.0,ds004259:1.0.0,ds004261:2.0.0,ds004280:1.0.1,ds004283:1.0.3,ds004285:1.0.0,ds004286:1.0.0,ds004299:1.0.0,ds004301:1.0.2,ds004302:1.0.1,ds004312:1.0.3,ds004323:1.0.0,ds004327:1.0.3,ds004331:1.0.4,ds004332:1.3.0,ds004341:1.0.0,ds004346:1.0.8,ds004349:1.0.0,ds004359:1.0.0,ds004392:1.0.0,ds004393:1.0.4,ds004401:1.3.0,ds004406:1.0.0,ds004440:1.0.1,ds004443:1.0.0,ds004450:1.0.1,ds004455:1.1.0,ds004456:1.0.1,ds004458:1.0.2,ds004466:1.0.2,ds004467:1.0.0,ds004469:1.1.4,ds004470:1.0.1,ds004471:1.0.1,ds004473:1.0.2,ds004475:1.0.3,ds004478:1.0.2,ds004482:1.0.0,ds004484:1.0.1,ds004488:1.1.1,ds004496:2.1.2,ds004498:1.0.0,ds004499:1.0.3,ds004505:1.0.4,ds004513:1.0.4,ds004516:2.0.2,ds004529:1.1.1,ds004533:1.0.0,ds004539:1.1.1,ds004542:1.0.0,ds004544:1.0.0,ds004553:1.0.1,ds004556:1.0.1,ds004557:1.1.0,ds004560:1.0.1,ds004562:1.0.4,ds004564:1.0.1,ds004581:2.0.1,ds004590:1.0.0,ds004592:1.0.1,ds004594:1.0.1,ds004597:2.0.0,ds004604:2.0.0,ds004605:1.0.1,ds004611:1.0.2,ds004627:1.1.0,ds004630:1.1.3,ds004631:1.0.0,ds004636:1.0.4,ds004640:1.0.4,ds004645:1.0.0,ds004647:1.0.2,ds004648:1.0.0,ds004650:1.0.2,ds004654:1.0.1,ds004656:1.1.0,ds004662:1.1.0,ds004670:1.0.1,ds004692:1.0.0,ds004693:1.0.3,ds004697:1.0.2,ds004698:2.0.1,ds004710:1.1.0,ds004711:1.0.0,ds004712:2.0.1,ds004715:1.0.3,ds004717:1.0.0,ds004718:1.1.1,ds004720:1.0.0,ds004725:1.0.1,ds004730:1.0.0,ds004731:1.0.0,ds004733:1.0.0,ds004737:2.0.0,ds004743:1.0.0,ds004765:1.0.0,ds004775:1.1.1,ds004776:1.0.0,ds004783:1.0.1,ds004786:1.0.1,ds004787:1.1.0,ds004791:1.0.0,ds004795:1.0.0,ds004798:1.0.5,ds004808:1.0.0,ds004814:1.0.0,ds004815:1.0.1,ds004829:1.0.1,ds004835:1.0.0,ds004837:1.0.0,ds004848:1.0.1,ds004856:1.2.0,ds004866:1.0.0,ds004868:1.0.4,ds004884:1.0.2,ds004889:1.1.2,ds004892:1.0.1,ds004894:1.0.0,ds004909:1.1.0,ds004910:1.0.1,ds004917:1.0.1,ds004920:1.1.1,ds004928:1.0.0,ds004934:1.0.0,ds005754:1.1.0,ds004937:1.0.1,ds004945:1.0.0,ds004946:1.0.0,ds004956:1.0.1,ds004957:1.0.2,ds004958:1.0.0,ds004965:1.0.1,ds004996:2.0.0,ds005003:2.0.0,ds005009:1.0.0,ds005012:1.0.3,ds005016:1.1.1,ds005017:1.0.2,ds005026:1.0.0,ds005027:1.0.3,ds005038:1.0.3,ds005040:1.2.0,ds005047:1.0.7,ds005050:1.0.0,ds005056:1.0.0,ds005063:1.0.0,ds005069:1.0.0,ds005072:1.0.1,ds005073:1.0.0,ds005075:1.0.1,ds005085:1.0.0,ds005115:1.2.0,ds005118:1.0.0,ds005123:1.1.3,ds005125:1.0.0,ds005126:1.0.3,ds005127:1.0.4,ds005128:1.0.0,ds005134:1.0.0,ds005139:1.0.7,ds005148:1.1.0,ds005165:1.0.4,ds005169:1.0.0,ds005191:1.0.2,ds005194:1.1.0,ds005215:1.0.0,ds005216:1.1.0,ds005230:1.0.0,ds005234:2.1.7,ds005239:1.0.1,ds005250:1.1.5,ds005261:2.0.2,ds005263:1.0.0,ds005264:1.0.0,ds005265:1.0.0,ds005266:1.0.0,ds005267:1.0.1,ds005279:1.0.3,ds005295:1.0.2,ds005299:1.0.0,ds005304:1.0.3,ds005355:1.0.1,ds005357:1.0.0,ds005360:1.0.0,ds005364:1.0.0,ds005365:1.0.1,ds005366:2.0.0,ds005371:1.1.3,ds005381:1.0.0,ds005386:1.0.0,ds005401:1.1.1,ds005412:1.0.0,ds005415:1.0.0,ds005418:3.2.0,ds005422:1.0.1,ds005427:1.1.1,ds005449:1.0.0,ds005454:1.0.0,ds005455:1.1.5,ds005464:1.0.1,ds005469:2.0.0,ds005479:1.0.3,ds005492:1.0.0,ds005498:2.0.0,ds005504:1.0.0,ds005518:1.0.1,ds005529:1.0.1,ds005530:1.0.9,ds005531:1.0.0,ds005533:1.0.0,ds005559:1.0.1,ds005571:1.0.1,ds005573:1.0.0,ds005576:1.0.0,ds005581:1.0.0,ds005588:1.0.0,ds005595:1.0.0,ds005596:1.1.1,ds005597:1.0.1,ds005598:1.0.0,ds005600:1.1.0,ds005602:1.0.0,ds005603:1.0.1,ds005604:1.0.1,ds005619:1.1.0}}

\definecolor{iccvblue}{rgb}{0.21,0.49,0.74}

\def\paperID{1756} 
\def\confName{ICCV}
\def\confYear{2025}

\title{An OpenMind for 3D medical vision self-supervised learning}

\author{Tassilo Wald~\thanks{Equal contribution, Author order among the co-first authors may be adjusted for individual use.}~$^{,1,2,3}$,\quad Constantin Ulrich\footnotemark[1]$^{~,1,4,5}$,\quad Jonathan Suprijadi\footnotemark[1]~$^{,}$\thanks{Work done while at DKFZ.}, \\
Sebastian Ziegler$^{1,2}$,\quad Michal Nohel$^{6,7}$,\quad Robin Peretzke$^{1,4}$\\ Gregor Köhler\footnotemark[2]~$^{,1}$,\quad  Klaus Maier-Hein$^{1,2,3,4,5,8}$\\
\small $^{1}$ Division of Medical Image Computing, German Cancer Research Center (DKFZ), Heidelberg, Germany \\
  \small$^{2}$~Helmholtz Imaging, DKFZ; $^{3}$~Faculty of Mathematics and Computer Science, Heidelberg University\\
\small $^{4}$~Medical Faculty, Heidelberg University; \small$^{5}$~National Center for Tumor Diseases (NCT), Heidelberg\\
\small$^{6}$~Faculty of Electrical Engineering and Communication, Brno University of Technology, Czech Republic\\
\small$^{7}$~University hospital Ostrava, Department of Deputy director for science, research and education\\
    \small $^{8}$~Pattern Analysis and Learning Group, Department of Radiation Oncology \\ 
{\tt\small tassilo.wald@dkfz-heidelberg.de}
}
\begin{document}
\maketitle

\begin{abstract}
The field of self-supervised learning (SSL) for 3D medical images lacks consistency and standardization.
While many methods have been developed, it is impossible to identify the current state-of-the-art, due to i) varying and small pre-training datasets, ii) varying architectures, and iii) being evaluated on differing downstream datasets. 
In this paper, we bring clarity to this field and lay the foundation for further method advancements through three key contributions: We a) publish the largest publicly available pre-training dataset comprising \datasetsize~3D brain MRI volumes, enabling all practitioners to pre-train on a large-scale dataset. We b) benchmark existing 3D self-supervised learning methods on this dataset for a state-of-the-art CNN and Transformer architecture, clarifying the state of 3D SSL pre-training. Among many findings, we show that pre-trained methods can exceed a strong from-scratch nnU-Net ResEnc-L baseline. Lastly, we c) publish the code of our pre-training and fine-tuning frameworks and provide the pre-trained models created during the benchmarking process to facilitate rapid adoption and reproduction. Available \href{https://github.com/MIC-DKFZ/nnssl}{here}.

\end{abstract}

\section{Introduction}
\label{sec:intro}

Self-Supervised Learning (SSL) has emerged as a transformative approach in deep learning, enabling the extraction of robust and general representations from data-rich domains \citep{assran2023self, oquab2023dinov2, he2022masked}. This paradigm has led to significant advancements in areas with abundant data, such as natural language processing and natural image computer vision. 
In the domain of 3D medical imaging, where annotation is expensive and annotated data scarce, SSL pre-training holds great promise as it can learn transferable representations without requiring extensive manual labeling efforts. 
\noindent However, currently the field of medical image analysis follows two predominant approaches to develop vision models: training models from scratch, often using the nnU-Net framework \citep{isensee2021nnu}, or adopting supervised pre-training with large, monolithic annotated datasets \citep{wasserthal2023totalsegmentator, qu2024abdomenatlas,huang2023stu} or aggregates of many datasets to increase supervised data scale \citep{ulrich2023multitalent, liwell}. While the latter approach indicates a willingness to adopt pre-training paradigms, it raises the question why self-supervised learning is hardly pursued in 3D medical imaging. We believe this lack of SSL method development and adoption can be attributed to two major factors:

\noindent \textbf{1) A lack of a large, open-source 3D dataset for pre-training.} 
Acquiring large unlabeled datasets for self-supervised pre-training is a non-trivial task. While large studies like the UK-BioBank or the NIH Adolescent Brain Cognitive Development (ABCD) exist and hold $>$100k or $>$40k 3D Volumes respectively, access to them depends on a positive internal review process requiring an appropriate project proposal. To add to the complexity, datasets are often subject to unique Data Usage Agreements (DUAs) that impose various restrictions on recipients. These rules may include: (i) mandatory acknowledgments, (ii) internal administrative review of manuscripts prior to submission \citep{OASIS_DUA,PPMI_DUA,ADNIDUA,MajoDUA,HABS_HD_DUA}, (iii) inclusion of a consortium in the author list \citep{ADNIDUA, MajoDUA, HABS_HD_DUA}, or (iv) requiring the dataset’s name to appear in the manuscript title \citep{MajoDUA}. While acknowledgements are reasonable and exemptions for the remaining restrictions can be requested, these DUAs create legal barriers that hinder reproduction or creation of large-scale pre-training datasets. 
Consequently, the majority of current SSL methods are either developed on large, restricted or proprietary datasets \citep{wald2024revisiting, wang2023mis, munk2024amaes} or are developed on small-scale publicly available datasets \citep{chen2023masked,zhuang2023advancing, he2023geometric,zhou2021models,wu2024voco,tang2022self,tang2024hyspark}, leading to the current fragmented landscape of pre-training datasets. 

\noindent \textbf{2) A lack of comparability:}
To adopt existing pre-trained models or conduct one's own SSL pre-training, it is important to know which model or pre-training method is the best.
Currently, this is an impossible task, due to: i) Different pre-training dataset selection (as previously highlighted). ii) Different architecture choices. Some methods use CNNs for pre-training \citep{wald2024revisiting,zhou2021models, munk2024amaes}, some ViT-like Transformers \citep{chen2023masked, zhuang2023advancing}, others Swin Transformers \citep{tang2022self,wu2024voco} and even others hybrid architectures \citep{tang2024hyspark, wang2023mis}. iii)  Insufficient downstream dataset choices, potentially providing unreliable results. \\

\noindent
In this paper we address these issues and provide three key contributions: i) We share the largest publicly available 3D dataset, comprising \datasetsize~3D volumes of 23 different brain MRI modalities. ii) We conduct a comprehensive self-supervised learning benchmark leveraging this dataset to assess the performance of state-of-the-art CNN and transformer-based architectures on 15 segmentation and classification downstream datasets. iii) We open-source our pre-training and fine-tuning frameworks as well as the trained model checkpoints to enable rapid reproduction and novel method development in the field of 3D medical image segmentation.

\section{The OpenMind Dataset}
\label{sec:dataset}

\begin{figure}
    \centering
    \includegraphics[width=.97\linewidth]{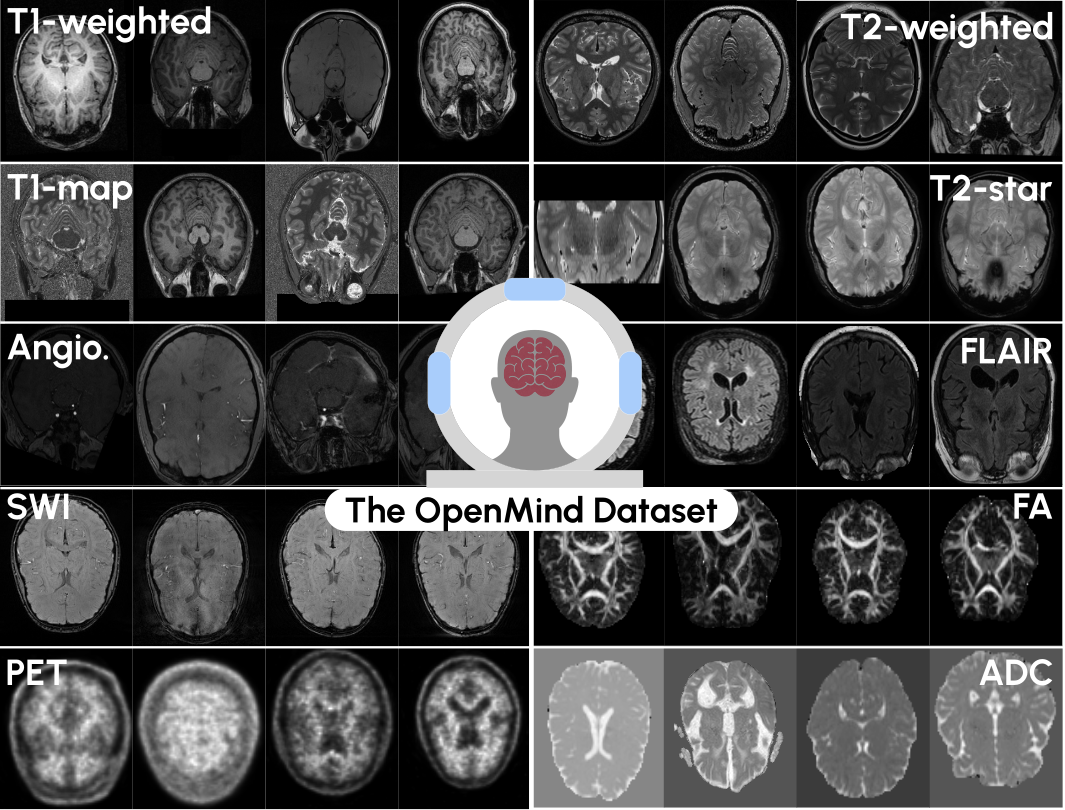}
    \caption{The OpenMind dataset contains \datasetsize~3D Head-And-Neck volumes of 23 different modalities. It represents the largest openly accessible dataset of 3D medical images currently available.}
    \label{fig:dataset_overview_figure}
\end{figure}

To provide practitioners access to a large-scale dataset for self-supervised learning, we introduce the OpenMind dataset: a large-scale, easily accessible resource containing \datasetsize~Head-and-Neck MRI volumes across various modalities, sourced and curated from the OpenNeuro platform. As the largest freely accessible 3D brain imaging dataset to date, OpenMind surpasses the NIH Adolescent Brain Cognitive Decline initiative~\citep{national2015adolescent} by a factor of three, and exceeds the recent CT-Rate dataset~\citep{hamamci2024developing} by a factor of two, while being released under a permissive CC-BY license (see \cref{tab:dataset_comparison}).

\begin{table}
    \centering
    \caption{Large medical datasets are often restricted through custom DUAs limiting their overall applicability.}
    \label{tab:dataset_comparison}
    \resizebox{.9\linewidth}{!}{%
    \begin{tabular}{lrrrr}
    \toprule
         Dataset & Patients & 3D Volumes & Modalities & Access\\
         \midrule
         ABCD~\citep{national2015adolescent} & 11,385 & 41,580 & 2 & Restricted \\
         PPMI~\citep{marek2018parkinson} & 2,011 & 9,554 & 5 & Restricted\\
         ADNI~\citep{mueller2005alzheimer} & 1,563 & 9,548 & 3 & Restricted \\
         OASIS3 & 1,376 & 15,872 & 7 & Restricted \\
         OASIS4 & 661 & 4,021 & 5 & Restricted \\
         BraTS21~\citep{baid2021rsna} & 1,251 & 5,004 & 4 & CC-BY-4.0\\ 
         CT-RATE~\cite{national2015adolescent} & 21,304 & 50,188 & 2 & CC-BY-NC-4.0 \\
         \midrule
         \textbf{OpenMind} & 34,191 & 113,921 & 24 & CC-BY-4.0\\ 
         \bottomrule
    \end{tabular}}
\end{table}

\subsection{Dataset creation}

All data was sourced or derived from datasets available on the OpenNeuro platform~\citep{markiewicz2021openneuro}. OpenNeuro (formerly known as openfMRI) is a publicly accessible data repository designed to support neurological research, adhering to the FAIR principles (Findable, Accessible, Interoperable, Reusable). The platform currently hosts over 1,200 public datasets, encompassing data from more than 51,300 subjects.
Due to the diverse nature of its collection, OpenNeuro includes a wide array of modalities, such as MRI, PET, fMRI, MEG, EEG, and iEEG, from studies involving both healthy and diseased subjects, enabling research across various aspects of neurology.
From this extensive collection, all available 3D MRIs, such as T1-weighted and T2-weighted scans, as well as 4D diffusion-weighted MRIs were collected from 800 different studies (citations [55-854]). This aggregation resulted in a total of 71k 3D MRI scans\footnote{This includes a minority of 653 PET scans.} and 15k 4D diffusion-weighted images from a total of 34,139 subjects. 
While existing 3D SSL methods could be directly trained on the 71k 3D MRI scans spanning multiple MRI sequences, the 4D diffusion-weighted images required pre-processing to create 3D derivatives which we describe in the following section. Overall, due to the high amount of independent studies, the OpenMind dataset provides great levels of diversity regarding participant age distribution, countries of origin, MRI modalities, scanner manufacturer, model and scanning protocols.    

\paragraph{DWI Preprocessing}
Diffusion-weighted imaging (DWI) is an advanced MRI modality that captures the random motion of water molecules in tissue by applying diffusion-sensitive gradient pulses, enabling the assessment of microstructural properties. It is widely used for detecting abnormalities such as strokes, tumors, and white matter changes.
In clinical and research settings, DWI data is often preprocessed into more interpretable 3D derivatives, including mean diffusion (MD) maps, and fractional anisotropy (FA) maps, which simplify analysis and interpretation. Additionally, T2-weighted images can be derived from DWI scans.
Consequently, we pre-processed the DWI images into MD maps, FA maps, and T2-weighted images, resulting in an additional 43k 3D images. The detailed pre-processing pipeline is described in \cref{apx:dwi_preprocessing}.

\begin{figure}
    \centering
    \includegraphics[width=\linewidth]{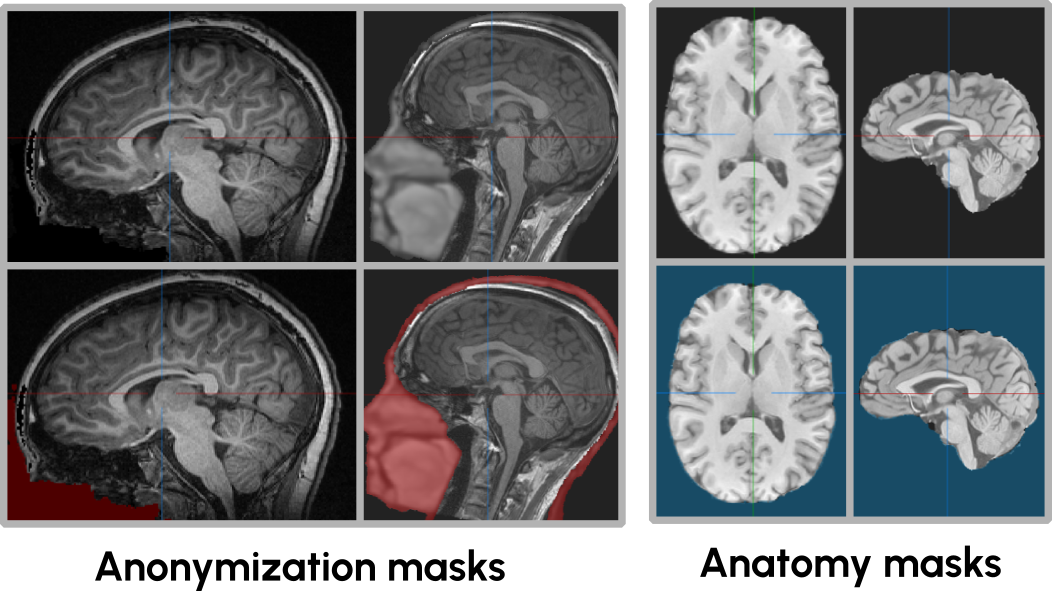}
    \caption{Head-and-Neck scans are often defaced, have the face blurred or have been brain-extracted to guarantee patient privacy. This can potentially harm reconstruction-based SSL methods. We provide anonymization and anatomy masks to allow taking this into account during method development.}
    \label{fig:deface_and_foreground}
\end{figure}

\paragraph{Defacing}
Protecting patient privacy is a big concern in the medical domain, hence, the facial region of Head-and-Neck scans are commonly anonymized. Common anonymization techniques include defacing -- replacing the volume with a constant value --, blurring everything but the brain or brain extraction, see \cref{fig:deface_and_foreground}. 
These augmented regions can interfere with reconstruction-based SSL methods, potentially penalizing models for attempting to reconstruct plausible anatomical features in anonymized regions.
Therefore, masks delineating these regions are needed to allow excluding them from the loss computation.
Unfortunately, not all datasets provide these masks, despite applying some sort of anonymization.
To address this issue, we generate associated \textit{anonymization masks} and \textit{anatomy masks} for cases where these are not provided (\cref{fig:deface_and_foreground} bottom). The anonymization mask identifies areas where anatomical structures have been artificially modified or removed and where no loss should be calculated, whereas anatomy masks indicate where relevant foreground is present, allowing to avoid sampling non-empty patches which may speed up the convergence of the pre-training methods. Both of these masks are created using the model proposed by \citet{youknow}.

\paragraph{Meta-Data enrichment}
While OpenNeuro follows the Brain Imaging Data Structure (BIDS) format, metadata organization varies significantly across datasets. Participant demographics (e.g., age, sex, health status) and image acquisition details (e.g., repetition times, scanner types, imaging protocols) are inconsistently reported, creating challenges for systematic analysis. We curated and harmonized all available, disparate metadata into a unified, complementary metadata structure. This standardized framework ensures consistency and enables users to efficiently filter the OpenMind dataset based on specific selection criteria, such as subject characteristics or imaging parameters, enhancing the flexibility of the dataset and simplifying the extraction of relevant sub-populations.\\
\noindent In addition to the harmonized metadata, we provide an \textit{Image Quality Score (IQS)} for each imaging modality of each of the 800 datasets. This score represents a subjective usability score of the images for self-supervised pre-training. The assessment considers factors such as image noise levels, motion artifacts, and other image quality-degrading elements, condensing these evaluations into a unified score ranging from 1 to 5 with 1 representing the highest quality images. The quality scores were determined through manual inspection of multiple images from each modality in each dataset and were assessed independently by two raters to ensure reliability  (more details in \cref{apx:image_quality_score}). 
Together, the harmonized metadata and Image Quality Score are introduced to test the effects of simple data-filtering techniques on self-supervised methods trained on the OpenMind dataset. If significant effects of pre-training data choice can be measured, the OpenMind datasets would show to have sufficient size to enable data-centric research in the domain of self-supervised learning for 3D medical image analysis, akin to~\citep{gadre2023datacomp,abbas2023semdedup,oquab2023dinov2}.\\

\noindent\textbf{Useability}
To maximize the utility of the OpenMind dataset, we provide the final dataset of \datasetsize~3D images on \href{https://huggingface.co/datasets/AnonRes/OpenMind}{Hugging Face} together with anonymization and anatomy masks, as well as the harmonized metadata including the Image Quality Score.\\

\noindent\textbf{Why benchmarking matters}
While the OpenMind dataset addresses the issue of data accessibility and data scale in self-supervised learning for 3D medical imaging, the current state-of-the-art remains unknown. 
To establish a reference point and provide an initial set of baselines, we conduct a comprehensive benchmarking effort to compare existing 3D SSL methods within a common framework. 

\section{The OpenMind Benchmark}
The field of self-supervised learning (SSL) for 3D medical imaging remains fragmented due to inconsistencies in pre-training datasets, architectures, and downstream evaluations. Without standardized benchmarks, it is impossible to fairly compare methods and determine the most effective approaches \citep{wald2024revisiting}. The OpenMind Benchmark addresses this challenge by evaluating a broad range of SSL methods under controlled conditions, using a common pre-training dataset, standardized architectures, and diverse downstream tasks, which we detail in this section.

\subsection{Standardization}
To establish a standardized and fair evaluation, all self-supervised learning (SSL) methods trained in the benchmark were trained on the OpenMind dataset, ensuring that pre-training dataset variability does not influence results.
Further, we use two state-of-the-art architectures, the ResEnc-L Residual Encoder U-Net \citep{isensee2024nnu} to represent convolutional neural networks (CNNs) and the recent Primus-M Transformer~\citep{wald2025primus}, allowing us to analyze SSL effectiveness across different architectural paradigms.
Pre-trained models are then evaluated on a diverse set of downstream datasets for segmentation and classification tasks.\\

\noindent\textbf{Downstream Datasets}
All downstream datasets used are split 50/50 into a train and test set. Moreover, we group our datasets into development datasets and test datasets. 
The development datasets serve to optimize SSL method hyperparameters on their train split, should the method not be directly applicable on the OpenMind dataset with original hyperparameters. During this process, the most important method-specific hyperparameters as well as learning rate and weight decay are optimized based on an additional 80/20 split on the training set of the development datasets. Aside from this, the development datasets are used to determine the optimal fine-tuning schedules, hence results of the optimal fine-tuning schedules are reported on the validation split of the dataset (\cref{tab:finetuning_cnns} and \cref{tab:fine-tuning_transformer}). \\
To avoid potential biases introduced during the optimization process on the development datasets, we withheld additional test datasets, which we left untouched and only evaluated after finishing all method optimizations on the development datasets.
Moreover, since datasets tend to be small and targets diverse, results are inherently noisy and may not generalize from one dataset to another. To reduce the chance of over optimizing for one dataset, we include four development datasets as well as eight test segmentation datasets and three test classification datasets.\\

\paragraph{Development datasets:}
\begin{enumerate}
    \item \textit{Atlas R2.0 (ATL)} \citep{liew2022large}: Brain scans of stroke patients imaged through T1 weighted MRIs.
    \item \textit{Stanford Brain Metastasis (SBM)} \citep{grovik2020deep}: Brain scans of metastasis imaged through pre-contrast T1, contrast-enhanced gradient echo, contrast-enhanced spin echo and a T2 FLAIR sequence . 
    \item \textit{AMOS22} (AMO)~\citep{ji2022amos}: CT and MRI scans of 15 abdominal organs.
    \item \textit{KiTS19} (KIT)~\citep{heller2021state}: CT images of kidneys and kidney tumors.
\end{enumerate}
While AMOS and KiTS can be considered out-of-distribution due to the pre-training images exclusively depicting the head-and-neck region, these datasets were found to be robust benchmarking datasets in \citep{isensee2024nnu}, hence were included despite their shift. Further, during development, we only optimized the methods for segmentation, due to it being the most common task in 3D medical image analysis~\citep{wald2025primus}.\\
\\
\textbf{Test datasets (Segmentation):}
\begin{enumerate}
    \item \textit{ISLES (ISL)} ~\citep{hernandez2022isles}: DWI (b=1000) and ADC map brain MRI scans of patients with delineations of stroke lesions.  
    \item \textit{HNTS-MRG (HNT)} ~\citep{hntsmrg2024wahid}: MR images of patients with head-and-neck primary tumors and metastatic lymph-nodes pre and mid-treatment of which we use the pre-treatment images and labels.
    \item \textit{HaN-Seg OAR (HAN)} ~\citep{podobnik2023han}: CT and MR images of Head-and-Neck organs-at-risk of which we use only the MR images.
    \item \textit{MS FLAIR (MSF)} ~\citep{MUSLIM2022msflair}: FLAIR MRIs of hyperintense multiple-sclerosis lesions with associated delineations. 
    \item \textit{ToP-CoW (TPC)} ~\citep{topcowchallenge}: Time-of-Flight (ToF) Magnetic Resonance Angiography (MRA) of the Circle of Willis and associated annotations of the arterial structure. 
    \item \textit{Yale Brain Metastasis (YBM)} \citep{ramakrishnan2024large}: Pre and post-contrast T1, T2 and FLAIR MRI images of brain metastases. 
    \item \textit{Cosmos (COS)} ~\citep{cosmos22challenge}: Carotid vessel wall segmentations and atherosclerosis diagnosis, with contours of the vessel walls on 3D-VISTA MRI.
    \item \textit{ACDC (ACD)} ~\citep{bernard2018deep}: Cardiac cine-MRI image of three segmented ventricular structures.

\end{enumerate}

\noindent \textbf{Test datasets (Classification):}
\begin{enumerate}
    \item \textit{MR-Net (MRN)} \citep{bien2018mrnet}: Knee T1w, T2w and Proton Density (PD) weighted MRIs of ACL or meniscal tears as well as other abnormalities which we classify.
    \item \textit{RSNA-SpineFrac (RSN)} \citep{lin2023rsna}: CT Spine fractures with the goal of identifying lumbar spine degenerative conditions. We classify fracture and no-fracture.
    \item \textit{ABIDE (ABI)} \citep{di2014autism}: T1w MRI images of either healthy controls or subjects with autism, on which we conduct binary classification -- autism or not-autism.
\end{enumerate}

\paragraph{Pre-training configuration}
To ensure a fair comparison, all SSL methods are trained under comparable pre-training and equal fine-tuning schedules. Specifically, we pre-train all methods for a total of 1000 epochs with 250 steps per epoch each, with each method trained DDP on four 40GB A100s to decrease pre-training time.
Furthermore, the pre-trained models were fine-tuned for a total amount of 150 or 1000 epochs with 250 steps per epoch and a common batch size of 2 for both architectures following an identical learning rate schedule implemented within the \href{https://github.com/TaWald/nnUNet}{nnU-Net framework} \citep{isensee2021nnu} for the segmentation downstream datasets. For classification downstream datasets, all models were fine-tuned for 200 dataset epochs within the public \href{https://github.com/MIC-DKFZ/image_classification}{Image classification framework}, which supports 3D classification. We provide more detailed descriptions of the fine-tuning process and frameworks in \cref{apx:finetuning_details}.

\subsection{Benchmark scope}
In the benchmark, three key aspects of SSL are investigated, highlighting the possible research directions the OpenMind dataset and benchmark can enable:
\begin{enumerate}[label=\roman*)]
    \item \textbf{SSL method performance:} We compare the efficacy of various SSL methods in learning transferable representations to various downstream segmentation and classification datasets when using the same pre-training dataset, using two common architectures and a unified fine-tuning schedule.
    \item \textbf{Fine-tuning methodology} We evaluate various fine-tuning schedules for both architectures to investigate the importance of a well-configured downstream fine-tuning schedule.
    \item \textbf{Data-centric baselines} We evaluate the influence of data-filtering on the large OpenMind dataset, to evaluate if data-centric research can be conducted on this scale. 
\end{enumerate}

\noindent \textbf{SSL Methods}
Various SSL methods native to 3D medical images exist. In the benchmark, we include contrastive methods, like Volume Contrastive (VoCo)~\citep{wu2024voco}, a 3D SimCLR adaptation, the pseudo-supervised VolumeFusion (VF)~\citep{wang2023mis}, default masked autoencoders (MAE)~\citep{he2022masked}, sparse adaptations of them for CNNs (S3D)~\citep{wald2024revisiting}, masked image modeling (SimMIM)~\citep{chen2023masked} and, the denoising and masked auto-encoder combination ModelsGenesis (MG)~\citep{zhou2021models}. Moreover, we incorporate the SwinUNETR pre-training scheme (SwinUNETR)~\cite{tang2022self}, which integrates masked inpainting, contrastive learning and rotation prediction\footnote{We refer to the pre-training scheme as SwinUNETR, not to be confused with the identically named architecture.}.
We tried including the 3D geometric-matching-based pre-training method GVSL~\cite{he2023geometric}, but the method failed to converge both when using the original implementation and when reimplementing it within our framework, regardless of whether we trained it on our OpenMind dataset or the ABCD-NIH dataset. We provide a brief description of each included SSL method and its used hyperparameters in \cref{apx:method_optim}. 
\noindent Methods, which did not feature publicly available Code that could be tested for functionality or integrated into our repository, were excluded, e.g. GL-MAE~\citep{zhuang2023advancing}. Moreover, more recent self-supervised methods were given higher priority for inclusion in our benchmark than older methods.
Furthermore, prominent 2D methods were excluded, as adapting their configuration from the natural imaging domain to the 3D medical imaging domain would require substantial development effort, e.g. due to large batch size requirements leading to VRAM issues.

\paragraph{Fine-tuning Schedules}
Effective adaptation during fine-tuning is crucial for optimizing performance in medical image segmentation, as it determines how well a pre-trained model can transfer knowledge to a target dataset. We test five different fine-tuning strategies for adaptation to segmentation datasets, aiming to balance maintaining general pre-trained representations with adapting the encoder to specialized representations through varying learning rate (lr) schedules:
\begin{enumerate*}[label=\roman*)]
    \item nnU-Net’s \textit{default} training strategy employs a polynomial decay lr with an initial lr of 1e-2\footnote{Learning rate value of the ResEnc-L CNN only. Primus-M learning rate values are provided in \cref{apx:finetuning_details}}, which we reduce to 1e-3 for the transfer learning setting.
    \item The \textit{Frozen} schedule follows the same lr schedule but restricts training to the decoder.
    \item The \textit{Warm-Up} schedule introduces an initial phase with a linear increase in the lr before transitioning to the default schedule.
    \item The \textit{Valley} scheme prioritizes decoder adaptation by first training it with a linear decreasing lr, followed by a linear warm-up phase for the full network, and then the default schedule.
    \item Finally, the \textit{Sawtooth} scheme employs a two-stage warm-up: first for the decoder while keeping the encoder frozen, and then for the whole network, both with a linear increasing lr, before transitioning to the default schedule.
\end{enumerate*} \\
\noindent We train the majority of fine-tuning runs for 150 epochs to reduce computational cost. For these, each warm-up stage lasts 15 epochs. For the CNNs we provide additional fine-tuning experiments trained for 1000 epochs, in which the warm-up stage is extended to 50 epochs. For a more detailed description and visualization of this, we defer to \cref{apx:finetuning_details}, where we also provide additional information on the classification fine-tuning.

\paragraph{Dataset Filters}
When training on a large enough quantity of data, optimizing the dataset becomes an important aspect that can greatly improve the resulting model performance~\citep{oquab2023dinov2,gadre2023datacomp,abbas2023semdedup}. Given the scale and diversity of the OpenMind dataset, we evaluate four simple filtering schemes to measure if we are at a scale where data filtering can be effective. The filtering schemes implemented are: \begin{enumerate*}
    \item Filtering by the manually created Image Quality Score (IQS) at three thresholds, and
    \item filtering all modalities but T1w, T2w and FLAIR images, the most common downstream dataset modalities.
\end{enumerate*}

\paragraph{Reproducibility}
While evaluating this benchmark, a lot of pre-trained models were generated. To maximize reproducibility, we provide the checkpoints of all pre-training methods for both architectures. Further, we provide an integration into the nnU-Net framework that holds the provided fine-tuning code, enabling users to quickly adapt the pre-trained checkpoints to a desired downstream task.

\begin{table*}
    \centering
    \caption{\textbf{Segmentation results of tested pre-training methods.} CNN and Transformer architectures were pre-trained on the same dataset for the same amount of steps. All methods were fine-tuned with the optimal architecture specific fine-tuning scheme for 150 epochs, unless indicated by suffix \textit{1k}. Color ranges are set per-dataset and per-architecture. S3D and SimMIM are only shown for CNN or Transformer as they are architecture specific. \textit{nnU-Net def.: Dynamically planned default nnU-Net trained from scratch; 1k: Fine-tuned for 1000 epochs instead of 150 epochs; ID Mean: In distribution mean; OOD Mean: Mean across datasets of non Head-and-Neck datasets.}}
    \label{tab:segmentation_results}
    \resizebox{.95\linewidth}{!}{\begin{tabular}{l|rrrrrrrrr|rrr|rrr}
\toprule
& \multicolumn{15}{c}{Dice Similarity Coefficient (DSC) [\%] on ...}\\
\cline{2-16}
& \multicolumn{9}{c|}{Dataset of same anatomical region (ID)} & \multicolumn{3}{c|}{Dataset of OOD region} & \multicolumn{3}{c}{Average across ...} \\ 
PT Method & ATL & SBM & ISL & HNT & HAN & MSF & TPC & YBM & COS & ACD & AMO & KIT & ID  & OOD  & All \\
\midrule
nnU-Net def. 1k & \colorrangeATLcnnbench{58.70} & \colorrangeSBMcnnbench{59.98} & \colorrangeISLcnnbench{78.40} & \colorrangeHNTcnnbench{62.98} & \colorrangeHANcnnbench{53.37} & \colorrangeMSFcnnbench{52.19} & \colorrangeTPCcnnbench{79.50} & \colorrangeYBMcnnbench{58.43} & \colorrangeCOScnnbench{46.19} & \colorrangeACDcnnbench{91.10} & \colorrangeAMOcnnbench{88.00} & \colorrangeKITcnnbench{87.21} & \colorrangeIDMeancnnbench{61.08} & \colorrangeOODMeancnnbench{88.77} & \colorrangeMeancnnbench{68.00} \\
nnU-Net def. & \colorrangeATLcnnbench{56.08} & \colorrangeSBMcnnbench{60.41} & \colorrangeISLcnnbench{78.22} & \colorrangeHNTcnnbench{59.37} & \colorrangeHANcnnbench{32.27} & \colorrangeMSFcnnbench{55.84} & \colorrangeTPCcnnbench{76.78} & \colorrangeYBMcnnbench{56.73} & \colorrangeCOScnnbench{49.31} & \colorrangeACDcnnbench{90.72} & \colorrangeAMOcnnbench{83.88} & \colorrangeKITcnnbench{77.61} & \colorrangeIDMeancnnbench{58.33} & \colorrangeOODMeancnnbench{84.07} & \colorrangeMeancnnbench{64.77} \\\midrule
\multicolumn{16}{c}{ResEnc-L (CNN)}\\\midrule
Scratch 1k & \colorrangeATLcnnbench{58.21} & \colorrangeSBMcnnbench{53.43} & \colorrangeISLcnnbench{79.14} & \colorrangeHNTcnnbench{65.75} & \colorrangeHANcnnbench{58.24} & \colorrangeMSFcnnbench{54.90} & \colorrangeTPCcnnbench{79.94} & \colorrangeYBMcnnbench{56.12} & \colorrangeCOScnnbench{71.57} & \colorrangeACDcnnbench{92.09} & \colorrangeAMOcnnbench{88.73} & \colorrangeKITcnnbench{87.48} & \colorrangeIDMeancnnbench{64.15} & \colorrangeOODMeancnnbench{89.43} & \colorrangeMeancnnbench{70.47} \\
Scratch & \colorrangeATLcnnbench{57.02} & \colorrangeSBMcnnbench{54.29} & \colorrangeISLcnnbench{78.09} & \colorrangeHNTcnnbench{63.30} & \colorrangeHANcnnbench{56.11} & \colorrangeMSFcnnbench{55.47} & \colorrangeTPCcnnbench{76.18} & \colorrangeYBMcnnbench{54.42} & \colorrangeCOScnnbench{65.20} & \colorrangeACDcnnbench{91.97} & \colorrangeAMOcnnbench{85.24} & \colorrangeKITcnnbench{84.03} & \colorrangeIDMeancnnbench{62.23} & \colorrangeOODMeancnnbench{87.08} & \colorrangeMeancnnbench{68.44} \\
VoCo & \colorrangeATLcnnbench{57.14} & \colorrangeSBMcnnbench{59.62} & \colorrangeISLcnnbench{77.50} & \colorrangeHNTcnnbench{63.48} & \colorrangeHANcnnbench{51.12} & \colorrangeMSFcnnbench{54.90} & \colorrangeTPCcnnbench{75.12} & \colorrangeYBMcnnbench{56.92} & \colorrangeCOScnnbench{63.49} & \colorrangeACDcnnbench{91.44} & \colorrangeAMOcnnbench{85.60} & \colorrangeKITcnnbench{85.71} & \colorrangeIDMeancnnbench{62.14} & \colorrangeOODMeancnnbench{87.58} & \colorrangeMeancnnbench{68.50} \\
SwinUNETR & \colorrangeATLcnnbench{56.07} & \colorrangeSBMcnnbench{57.25} & \colorrangeISLcnnbench{77.45} & \colorrangeHNTcnnbench{61.64} & \colorrangeHANcnnbench{49.42} & \colorrangeMSFcnnbench{54.82} & \colorrangeTPCcnnbench{74.69} & \colorrangeYBMcnnbench{57.05} & \colorrangeCOScnnbench{65.68} & \colorrangeACDcnnbench{90.53} & \colorrangeAMOcnnbench{84.95} & \colorrangeKITcnnbench{85.54} & \colorrangeIDMeancnnbench{61.56} & \colorrangeOODMeancnnbench{87.01} & \colorrangeMeancnnbench{67.92} \\
SimCLR & \colorrangeATLcnnbench{57.15} & \colorrangeSBMcnnbench{59.72} & \colorrangeISLcnnbench{78.01} & \colorrangeHNTcnnbench{63.32} & \colorrangeHANcnnbench{51.56} & \colorrangeMSFcnnbench{55.68} & \colorrangeTPCcnnbench{77.77} & \colorrangeYBMcnnbench{59.14} & \colorrangeCOScnnbench{68.20} & \colorrangeACDcnnbench{91.76} & \colorrangeAMOcnnbench{86.06} & \colorrangeKITcnnbench{84.85} & \colorrangeIDMeancnnbench{63.40} & \colorrangeOODMeancnnbench{87.56} & \colorrangeMeancnnbench{69.44} \\
VF & \colorrangeATLcnnbench{57.42} & \colorrangeSBMcnnbench{59.88} & \colorrangeISLcnnbench{78.18} & \colorrangeHNTcnnbench{64.32} & \colorrangeHANcnnbench{51.67} & \colorrangeMSFcnnbench{57.42} & \colorrangeTPCcnnbench{76.11} & \colorrangeYBMcnnbench{59.31} & \colorrangeCOScnnbench{63.98} & \colorrangeACDcnnbench{91.57} & \colorrangeAMOcnnbench{85.38} & \colorrangeKITcnnbench{86.21} & \colorrangeIDMeancnnbench{63.14} & \colorrangeOODMeancnnbench{87.72} & \colorrangeMeancnnbench{69.29} \\
MG & \colorrangeATLcnnbench{58.03} & \colorrangeSBMcnnbench{61.57} & \colorrangeISLcnnbench{77.58} & \colorrangeHNTcnnbench{65.11} & \colorrangeHANcnnbench{54.69} & \colorrangeMSFcnnbench{55.25} & \colorrangeTPCcnnbench{77.14} & \colorrangeYBMcnnbench{58.67} & \colorrangeCOScnnbench{71.27} & \colorrangeACDcnnbench{91.74} & \colorrangeAMOcnnbench{86.35} & \colorrangeKITcnnbench{86.17} & \colorrangeIDMeancnnbench{64.37} & \colorrangeOODMeancnnbench{88.09} & \colorrangeMeancnnbench{70.30} \\
MAE & \colorrangeATLcnnbench{58.25} & \colorrangeSBMcnnbench{62.41} & \colorrangeISLcnnbench{77.89} & \colorrangeHNTcnnbench{66.58} & \colorrangeHANcnnbench{55.14} & \colorrangeMSFcnnbench{56.84} & \colorrangeTPCcnnbench{77.96} & \colorrangeYBMcnnbench{60.07} & \colorrangeCOScnnbench{70.85} & \colorrangeACDcnnbench{91.98} & \colorrangeAMOcnnbench{86.78} & \colorrangeKITcnnbench{86.12} & \colorrangeIDMeancnnbench{65.11} & \colorrangeOODMeancnnbench{88.30} & \colorrangeMeancnnbench{70.91} \\
S3D & \colorrangeATLcnnbench{58.76} & \colorrangeSBMcnnbench{64.09} & \colorrangeISLcnnbench{78.05} & \colorrangeHNTcnnbench{65.74} & \colorrangeHANcnnbench{52.81} & \colorrangeMSFcnnbench{56.08} & \colorrangeTPCcnnbench{78.81} & \colorrangeYBMcnnbench{59.18} & \colorrangeCOScnnbench{66.66} & \colorrangeACDcnnbench{92.01} & \colorrangeAMOcnnbench{86.16} & \colorrangeKITcnnbench{86.01} & \colorrangeIDMeancnnbench{64.46} & \colorrangeOODMeancnnbench{88.06} & \colorrangeMeancnnbench{70.36} \\

\midrule
\multicolumn{16}{c}{Primus-M (Transformer)}\\
\midrule
Scratch 1k & \colorrangeATLtransbench{56.77} & \colorrangeSBMtransbench{48.50} & \colorrangeISLtransbench{76.59} & \colorrangeHNTtransbench{58.40} & \colorrangeHANtransbench{53.40} & \colorrangeMSFtransbench{53.27} & \colorrangeTPCtransbench{76.32} & \colorrangeYBMtransbench{52.53} & \colorrangeCOStransbench{64.68} & \colorrangeACDtransbench{90.89} & \colorrangeAMOtransbench{87.24} & \colorrangeKITtransbench{85.57} & \colorrangeIDMeantransbench{60.05} & \colorrangeOODMeantransbench{87.90} & \colorrangeMeantransbench{67.01} \\
Scratch & \colorrangeATLtransbench{51.51} & \colorrangeSBMtransbench{43.26} & \colorrangeISLtransbench{75.23} & \colorrangeHNTtransbench{55.30} & \colorrangeHANtransbench{50.60} & \colorrangeMSFtransbench{54.00} & \colorrangeTPCtransbench{73.31} & \colorrangeYBMtransbench{50.30} & \colorrangeCOStransbench{62.11} & \colorrangeACDtransbench{90.93} & \colorrangeAMOtransbench{80.17} & \colorrangeKITtransbench{76.73} & \colorrangeIDMeantransbench{57.29} & \colorrangeOODMeantransbench{82.61} & \colorrangeMeantransbench{63.62} \\
VoCo & \colorrangeATLtransbench{46.80} & \colorrangeSBMtransbench{34.15} & \colorrangeISLtransbench{73.29} & \colorrangeHNTtransbench{51.06} & \colorrangeHANtransbench{47.64} & \colorrangeMSFtransbench{52.64} & \colorrangeTPCtransbench{65.52} & \colorrangeYBMtransbench{44.75} & \colorrangeCOStransbench{52.16} & \colorrangeACDtransbench{87.26} & \colorrangeAMOtransbench{65.81} & \colorrangeKITtransbench{70.21} & \colorrangeIDMeantransbench{52.00} & \colorrangeOODMeantransbench{74.43} & \colorrangeMeantransbench{57.61} \\
SwinUNETR & \colorrangeATLtransbench{47.23} & \colorrangeSBMtransbench{36.31} & \colorrangeISLtransbench{73.84} & \colorrangeHNTtransbench{50.15} & \colorrangeHANtransbench{46.49} & \colorrangeMSFtransbench{52.80} & \colorrangeTPCtransbench{66.23} & \colorrangeYBMtransbench{44.49} & \colorrangeCOStransbench{54.82} & \colorrangeACDtransbench{87.92} & \colorrangeAMOtransbench{66.17} & \colorrangeKITtransbench{70.39} & \colorrangeIDMeantransbench{52.49} & \colorrangeOODMeantransbench{74.82} & \colorrangeMeantransbench{58.07} \\
SimCLR & \colorrangeATLtransbench{54.61} & \colorrangeSBMtransbench{42.62} & \colorrangeISLtransbench{75.43} & \colorrangeHNTtransbench{56.75} & \colorrangeHANtransbench{50.80} & \colorrangeMSFtransbench{53.59} & \colorrangeTPCtransbench{70.08} & \colorrangeYBMtransbench{48.36} & \colorrangeCOStransbench{58.20} & \colorrangeACDtransbench{89.97} & \colorrangeAMOtransbench{75.75} & \colorrangeKITtransbench{81.27} & \colorrangeIDMeantransbench{56.72} & \colorrangeOODMeantransbench{82.33} & \colorrangeMeantransbench{63.12} \\
VF & \colorrangeATLtransbench{58.62} & \colorrangeSBMtransbench{47.37} & \colorrangeISLtransbench{77.56} & \colorrangeHNTtransbench{62.37} & \colorrangeHANtransbench{56.18} & \colorrangeMSFtransbench{55.00} & \colorrangeTPCtransbench{74.98} & \colorrangeYBMtransbench{53.96} & \colorrangeCOStransbench{69.74} & \colorrangeACDtransbench{91.41} & \colorrangeAMOtransbench{84.95} & \colorrangeKITtransbench{86.17} & \colorrangeIDMeantransbench{61.75} & \colorrangeOODMeantransbench{87.51} & \colorrangeMeantransbench{68.19} \\
MG & \colorrangeATLtransbench{56.50} & \colorrangeSBMtransbench{47.34} & \colorrangeISLtransbench{76.76} & \colorrangeHNTtransbench{58.42} & \colorrangeHANtransbench{54.02} & \colorrangeMSFtransbench{54.67} & \colorrangeTPCtransbench{73.65} & \colorrangeYBMtransbench{49.77} & \colorrangeCOStransbench{60.74} & \colorrangeACDtransbench{90.89} & \colorrangeAMOtransbench{82.15} & \colorrangeKITtransbench{84.45} & \colorrangeIDMeantransbench{59.10} & \colorrangeOODMeantransbench{85.83} & \colorrangeMeantransbench{65.78} \\
MAE & \colorrangeATLtransbench{61.16} & \colorrangeSBMtransbench{56.67} & \colorrangeISLtransbench{77.12} & \colorrangeHNTtransbench{66.12} & \colorrangeHANtransbench{57.24} & \colorrangeMSFtransbench{56.02} & \colorrangeTPCtransbench{78.31} & \colorrangeYBMtransbench{54.35} & \colorrangeCOStransbench{72.02} & \colorrangeACDtransbench{92.16} & \colorrangeAMOtransbench{87.16} & \colorrangeKITtransbench{86.74} & \colorrangeIDMeantransbench{64.34} & \colorrangeOODMeantransbench{88.69} & \colorrangeMeantransbench{70.42} \\
SimMIM & \colorrangeATLtransbench{60.28} & \colorrangeSBMtransbench{51.68} & \colorrangeISLtransbench{77.53} & \colorrangeHNTtransbench{62.76} & \colorrangeHANtransbench{56.74} & \colorrangeMSFtransbench{55.91} & \colorrangeTPCtransbench{77.00} & \colorrangeYBMtransbench{52.90} & \colorrangeCOStransbench{70.87} & \colorrangeACDtransbench{91.98} & \colorrangeAMOtransbench{86.57} & \colorrangeKITtransbench{85.92} & \colorrangeIDMeantransbench{62.85} & \colorrangeOODMeantransbench{88.16} & \colorrangeMeantransbench{69.18} \\
\bottomrule
\end{tabular}
}
\end{table*}

\begin{table*}
    \centering
    \caption{\textbf{Classification results of tested pre-training methods.} CNN and Transformer architectures were pre-trained and fine-tuned identically on the three classification datasets. \textit{PT Method: pre-training method}}
    \label{tab:classification_results}
    \resizebox{.95\linewidth}{!}{
    \begin{tabular}{l|ccc|c|ccc|c||ccc|c|ccc|c}
\toprule
Architecture & \multicolumn{8}{c||}{ResEnc-L (CNN)} & \multicolumn{8}{c}{Primus-M (Transformer)} \\
Metric & \multicolumn{4}{c|}{Average Precision} & \multicolumn{4}{c||}{Balanced Accuracy} & \multicolumn{4}{c|}{Average Precision} & \multicolumn{4}{c}{Balanced Accuracy} \\
PT Method & MRN & RSN & ABI & Mean & MRN & RSN & ABI & Mean & MRN & RSN & ABI & Mean & MRN & RSN & ABI & Mean \\
\midrule
Scratch & \colorrangeMRNetcnnAP{66.33} & \colorrangeRSNASpinecnnAP{65.39} & \colorrangeABIDEcnnAP{57.18} & \colorrangeMeancnnAP{62.97} & \colorrangeMRNetcnnBalAcc{76.61} & \colorrangeRSNASpinecnnBalAcc{62.25} & \colorrangeABIDEcnnBalAcc{53.29} & \colorrangeMeancnnBalAcc{64.05} & \colorrangeMRNettrAP{60.58} & \colorrangeRSNASpinetrAP{58.91} & \colorrangeABIDEtrAP{52.87} & \colorrangeMeantrAP{57.45} & \colorrangeMRNettrBalAcc{75.49} & \colorrangeRSNASpinetrBalAcc{57.10} & \colorrangeABIDEtrBalAcc{52.05} & \colorrangeMeantrBalAcc{61.55} \\
VoCo & \colorrangeMRNetcnnAP{71.99} & \colorrangeRSNASpinecnnAP{70.33} & \colorrangeABIDEcnnAP{62.23} & \colorrangeMeancnnAP{68.18} & \colorrangeMRNetcnnBalAcc{80.35} & \colorrangeRSNASpinecnnBalAcc{65.57} & \colorrangeABIDEcnnBalAcc{60.18} & \colorrangeMeancnnBalAcc{68.70} & \colorrangeMRNettrAP{67.92} & \colorrangeRSNASpinetrAP{62.09} & \colorrangeABIDEtrAP{53.22} & \colorrangeMeantrAP{61.08} & \colorrangeMRNettrBalAcc{77.68} & \colorrangeRSNASpinetrBalAcc{61.64} & \colorrangeABIDEtrBalAcc{51.28} & \colorrangeMeantrBalAcc{63.53} \\
SwinUNETR & \colorrangeMRNetcnnAP{70.72} & \colorrangeRSNASpinecnnAP{70.37} & \colorrangeABIDEcnnAP{62.32} & \colorrangeMeancnnAP{67.80} & \colorrangeMRNetcnnBalAcc{79.25} & \colorrangeRSNASpinecnnBalAcc{66.60} & \colorrangeABIDEcnnBalAcc{59.88} & \colorrangeMeancnnBalAcc{68.58} & \colorrangeMRNettrAP{64.98} & \colorrangeRSNASpinetrAP{60.79} & \colorrangeABIDEtrAP{56.84} & \colorrangeMeantrAP{60.87} & \colorrangeMRNettrBalAcc{76.61} & \colorrangeRSNASpinetrBalAcc{59.22} & \colorrangeABIDEtrBalAcc{55.60} & \colorrangeMeantrBalAcc{63.81} \\
SimCLR & \colorrangeMRNetcnnAP{68.74} & \colorrangeRSNASpinecnnAP{73.02} & \colorrangeABIDEcnnAP{59.17} & \colorrangeMeancnnAP{66.98} & \colorrangeMRNetcnnBalAcc{78.27} & \colorrangeRSNASpinecnnBalAcc{67.96} & \colorrangeABIDEcnnBalAcc{57.66} & \colorrangeMeancnnBalAcc{67.96} & \colorrangeMRNettrAP{67.67} & \colorrangeRSNASpinetrAP{60.26} & \colorrangeABIDEtrAP{55.59} & \colorrangeMeantrAP{61.18} & \colorrangeMRNettrBalAcc{76.69} & \colorrangeRSNASpinetrBalAcc{59.52} & \colorrangeABIDEtrBalAcc{54.15} & \colorrangeMeantrBalAcc{63.45} \\
VF & \colorrangeMRNetcnnAP{68.46} & \colorrangeRSNASpinecnnAP{66.22} & \colorrangeABIDEcnnAP{58.01} & \colorrangeMeancnnAP{64.23} & \colorrangeMRNetcnnBalAcc{78.43} & \colorrangeRSNASpinecnnBalAcc{63.28} & \colorrangeABIDEcnnBalAcc{55.60} & \colorrangeMeancnnBalAcc{65.77} & \colorrangeMRNettrAP{66.41} & \colorrangeRSNASpinetrAP{61.66} & \colorrangeABIDEtrAP{58.37} & \colorrangeMeantrAP{62.14} & \colorrangeMRNettrBalAcc{76.67} & \colorrangeRSNASpinetrBalAcc{60.14} & \colorrangeABIDEtrBalAcc{56.06} & \colorrangeMeantrBalAcc{64.29} \\
MG & \colorrangeMRNetcnnAP{68.85} & \colorrangeRSNASpinecnnAP{71.21} & \colorrangeABIDEcnnAP{59.69} & \colorrangeMeancnnAP{66.59} & \colorrangeMRNetcnnBalAcc{79.09} & \colorrangeRSNASpinecnnBalAcc{66.14} & \colorrangeABIDEcnnBalAcc{57.24} & \colorrangeMeancnnBalAcc{67.49} & \colorrangeMRNettrAP{68.86} & \colorrangeRSNASpinetrAP{61.24} & \colorrangeABIDEtrAP{52.41} & \colorrangeMeantrAP{60.84} & \colorrangeMRNettrBalAcc{77.73} & \colorrangeRSNASpinetrBalAcc{59.97} & \colorrangeABIDEtrBalAcc{54.13} & \colorrangeMeantrBalAcc{63.94} \\
MAE & \colorrangeMRNetcnnAP{64.53} & \colorrangeRSNASpinecnnAP{67.86} & \colorrangeABIDEcnnAP{58.31} & \colorrangeMeancnnAP{63.57} & \colorrangeMRNetcnnBalAcc{76.72} & \colorrangeRSNASpinecnnBalAcc{64.21} & \colorrangeABIDEcnnBalAcc{56.23} & \colorrangeMeancnnBalAcc{65.72} & \colorrangeMRNettrAP{67.00} & \colorrangeRSNASpinetrAP{60.28} & \colorrangeABIDEtrAP{53.09} & \colorrangeMeantrAP{60.13} & \colorrangeMRNettrBalAcc{77.26} & \colorrangeRSNASpinetrBalAcc{58.83} & \colorrangeABIDEtrBalAcc{51.84} & \colorrangeMeantrBalAcc{62.64} \\
S3D & \colorrangeMRNetcnnAP{70.08} & \colorrangeRSNASpinecnnAP{69.25} & \colorrangeABIDEcnnAP{60.65} & \colorrangeMeancnnAP{66.66} & \colorrangeMRNetcnnBalAcc{78.56} & \colorrangeRSNASpinecnnBalAcc{64.20} & \colorrangeABIDEcnnBalAcc{56.41} & \colorrangeMeancnnBalAcc{66.39} & \cellcolor{lightgray!30} - & \cellcolor{lightgray!30} - & \cellcolor{lightgray!30} - & \cellcolor{lightgray!30} - & \cellcolor{lightgray!30} - & \cellcolor{lightgray!30} - & \cellcolor{lightgray!30} - & \cellcolor{lightgray!30} - \\
SimMIM & \cellcolor{lightgray!30} - & \cellcolor{lightgray!30} - & \cellcolor{lightgray!30} - & \cellcolor{lightgray!30} - & \cellcolor{lightgray!30} - & \cellcolor{lightgray!30} - & \cellcolor{lightgray!30} - & \cellcolor{lightgray!30} - & \colorrangeMRNettrAP{65.27} & \colorrangeRSNASpinetrAP{59.62} & \colorrangeABIDEtrAP{53.94} & \colorrangeMeantrAP{59.61} & \colorrangeMRNettrBalAcc{77.15} & \colorrangeRSNASpinetrBalAcc{58.49} & \colorrangeABIDEtrBalAcc{53.93} & \colorrangeMeantrBalAcc{63.19} \\
\bottomrule
\end{tabular}
}
\end{table*}

\section{Results}
\label{sec:results}
In this section we highlight the most important insights of our benchmark, but not all due to spatial constraints. A comprehensive list of them is provided in \cref{apx:additional_benchmark_results}.


\begin{table}
    \centering
    \caption{\textbf{Effects of different CNN fine-tuning schedules.}}
    \label{tab:finetuning_cnns}
    \resizebox{.95\linewidth}{!}{
    \begin{tabular}{ll|lllllll|l}
\toprule
 & PT Method & VoCo & SwinUNETR & SimCLR & VF & MG & MAE & S3D & Average \\
Dataset & FT Schedule &  &  &  &  &  &  &  &  \\
\midrule
\multirow{5}{*}{SBM} & Default & \colorrangeSBMVoCocnn{73.86} & \colorrangeSBMSwinUNETRcnn{69.86} & \colorrangeSBMSimCLRcnn{72.79} & \colorrangeSBMVFcnn{72.57} & \colorrangeSBMMGcnn{73.37} & \colorrangeSBMMAEcnn{71.84} & \colorrangeSBMSDcnn{75.47} & \colorrangeSBMMeancnn{72.82} \\
 & Frozen & \colorrangeSBMVoCocnn{52.34} & \colorrangeSBMSwinUNETRcnn{38.59} & \colorrangeSBMSimCLRcnn{60.83} & \colorrangeSBMVFcnn{28.54} & \colorrangeSBMMGcnn{51.35} & \colorrangeSBMMAEcnn{59.45} & \colorrangeSBMSDcnn{56.55} & \colorrangeSBMMeancnn{49.67} \\
 & Warm-Up & \colorrangeSBMVoCocnn{70.87} & \colorrangeSBMSwinUNETRcnn{69.16} & \colorrangeSBMSimCLRcnn{73.94} & \colorrangeSBMVFcnn{73.11} & \colorrangeSBMMGcnn{74.67} & \colorrangeSBMMAEcnn{74.51} & \colorrangeSBMSDcnn{73.54} & \colorrangeSBMMeancnn{72.83} \\
 & Valley & \colorrangeSBMVoCocnn{70.26} & \colorrangeSBMSwinUNETRcnn{70.29} & \colorrangeSBMSimCLRcnn{72.82} & \colorrangeSBMVFcnn{75.81} & \colorrangeSBMMGcnn{74.68} & \colorrangeSBMMAEcnn{73.71} & \colorrangeSBMSDcnn{72.40} & \colorrangeSBMMeancnn{72.85} \\
 & Sawtooth & \colorrangeSBMVoCocnn{71.52} & \colorrangeSBMSwinUNETRcnn{71.05} & \colorrangeSBMSimCLRcnn{73.72} & \colorrangeSBMVFcnn{73.78} & \colorrangeSBMMGcnn{73.80} & \colorrangeSBMMAEcnn{74.79} & \colorrangeSBMSDcnn{73.97} & \colorrangeSBMMeancnn{73.23} \\\midrule
\multirow{5}{*}{ATL} & Default & \colorrangeATLVoCocnn{62.78} & \colorrangeATLSwinUNETRcnn{60.82} & \colorrangeATLSimCLRcnn{56.80} & \colorrangeATLVFcnn{63.83} & \colorrangeATLMGcnn{62.62} & \colorrangeATLMAEcnn{62.78} & \colorrangeATLSDcnn{62.16} & \colorrangeATLMeancnn{61.68} \\
 & Frozen & \colorrangeATLVoCocnn{58.27} & \colorrangeATLSwinUNETRcnn{54.27} & \colorrangeATLSimCLRcnn{57.95} & \colorrangeATLVFcnn{43.33} & \colorrangeATLMGcnn{57.27} & \colorrangeATLMAEcnn{54.49} & \colorrangeATLSDcnn{57.15} & \colorrangeATLMeancnn{54.68} \\
 & Warm-Up & \colorrangeATLVoCocnn{59.32} & \colorrangeATLSwinUNETRcnn{60.23} & \colorrangeATLSimCLRcnn{62.36} & \colorrangeATLVFcnn{63.20} & \colorrangeATLMGcnn{63.46} & \colorrangeATLMAEcnn{63.61} & \colorrangeATLSDcnn{62.34} & \colorrangeATLMeancnn{62.07} \\
 & Valley & \colorrangeATLVoCocnn{59.43} & \colorrangeATLSwinUNETRcnn{60.34} & \colorrangeATLSimCLRcnn{61.21} & \colorrangeATLVFcnn{64.00} & \colorrangeATLMGcnn{61.23} & \colorrangeATLMAEcnn{62.98} & \colorrangeATLSDcnn{63.38} & \colorrangeATLMeancnn{61.80} \\
 & Sawtooth & \colorrangeATLVoCocnn{60.42} & \colorrangeATLSwinUNETRcnn{61.17} & \colorrangeATLSimCLRcnn{61.34} & \colorrangeATLVFcnn{65.07} & \colorrangeATLMGcnn{61.54} & \colorrangeATLMAEcnn{63.47} & \colorrangeATLSDcnn{63.76} & \colorrangeATLMeancnn{62.40} \\\midrule

\multirow{5}{*}{AMO} & Default & \colorrangeAMOVoCocnn{85.93} & \colorrangeAMOSwinUNETRcnn{85.42} & \colorrangeAMOSimCLRcnn{87.12} & \colorrangeAMOVFcnn{86.05} & \colorrangeAMOMGcnn{86.93} & \colorrangeAMOMAEcnn{86.78} & \colorrangeAMOSDcnn{86.74} & \colorrangeAMOMeancnn{86.42} \\
 & Frozen & \colorrangeAMOVoCocnn{53.18} & \colorrangeAMOSwinUNETRcnn{61.26} & \colorrangeAMOSimCLRcnn{62.92} & \colorrangeAMOVFcnn{30.97} & \colorrangeAMOMGcnn{57.89} & \colorrangeAMOMAEcnn{50.83} & \colorrangeAMOSDcnn{52.48} & \colorrangeAMOMeancnn{52.79} \\
 & Warm-Up & \colorrangeAMOVoCocnn{86.98} & \colorrangeAMOSwinUNETRcnn{86.61} & \colorrangeAMOSimCLRcnn{87.10} & \colorrangeAMOVFcnn{86.58} & \colorrangeAMOMGcnn{87.34} & \colorrangeAMOMAEcnn{87.62} & \colorrangeAMOSDcnn{87.54} & \colorrangeAMOMeancnn{87.11} \\
 & Valley & \colorrangeAMOVoCocnn{85.91} & \colorrangeAMOSwinUNETRcnn{85.46} & \colorrangeAMOSimCLRcnn{86.72} & \colorrangeAMOVFcnn{86.55} & \colorrangeAMOMGcnn{87.09} & \colorrangeAMOMAEcnn{87.37} & \colorrangeAMOSDcnn{86.92} & \colorrangeAMOMeancnn{86.57} \\
 & Sawtooth & \colorrangeAMOVoCocnn{86.78} & \colorrangeAMOSwinUNETRcnn{85.76} & \colorrangeAMOSimCLRcnn{86.75} & \colorrangeAMOVFcnn{86.46} & \colorrangeAMOMGcnn{87.15} & \colorrangeAMOMAEcnn{87.48} & \colorrangeAMOSDcnn{86.45} & \colorrangeAMOMeancnn{86.69} \\\midrule

\multirow{5}{*}{KIT} & Default & \colorrangeKITVoCocnn{85.80} & \colorrangeKITSwinUNETRcnn{83.83} & \colorrangeKITSimCLRcnn{81.20} & \colorrangeKITVFcnn{84.34} & \colorrangeKITMGcnn{83.69} & \colorrangeKITMAEcnn{84.56} & \colorrangeKITSDcnn{82.21} & \colorrangeKITMeancnn{83.66} \\
 & Frozen & \colorrangeKITVoCocnn{69.33} & \colorrangeKITSwinUNETRcnn{69.59} & \colorrangeKITSimCLRcnn{65.22} & \colorrangeKITVFcnn{36.06} & \colorrangeKITMGcnn{71.39} & \colorrangeKITMAEcnn{63.97} & \colorrangeKITSDcnn{71.41} & \colorrangeKITMeancnn{63.85} \\
 & Warm-Up & \colorrangeKITVoCocnn{84.71} & \colorrangeKITSwinUNETRcnn{81.77} & \colorrangeKITSimCLRcnn{81.80} & \colorrangeKITVFcnn{83.51} & \colorrangeKITMGcnn{83.99} & \colorrangeKITMAEcnn{85.24} & \colorrangeKITSDcnn{82.20} & \colorrangeKITMeancnn{83.32} \\
 & Valley & \colorrangeKITVoCocnn{84.08} & \colorrangeKITSwinUNETRcnn{82.41} & \colorrangeKITSimCLRcnn{81.55} & \colorrangeKITVFcnn{85.00} & \colorrangeKITMGcnn{83.48} & \colorrangeKITMAEcnn{83.48} & \colorrangeKITSDcnn{83.78} & \colorrangeKITMeancnn{83.40} \\
 & Sawtooth & \colorrangeKITVoCocnn{85.12} & \colorrangeKITSwinUNETRcnn{83.26} & \colorrangeKITSimCLRcnn{83.10} & \colorrangeKITVFcnn{85.41} & \colorrangeKITMGcnn{84.15} & \colorrangeKITMAEcnn{84.28} & \colorrangeKITSDcnn{83.31} & \colorrangeKITMeancnn{84.09} \\\midrule
\multirow{5}{*}{Average} & Default & \colorrangeAVGVoCocnn{77.09} & \colorrangeAVGSwinUNETRcnn{74.98} & \colorrangeAVGSimCLRcnn{74.48} & \colorrangeAVGVFcnn{76.70} & \colorrangeAVGMGcnn{76.65} & \colorrangeAVGMAEcnn{76.49} & \colorrangeAVGSDcnn{76.65} & \colorrangeAVGMeancnn{76.15} \\
 & Frozen & \colorrangeAVGVoCocnn{58.28} & \colorrangeAVGSwinUNETRcnn{55.93} & \colorrangeAVGSimCLRcnn{61.73} & \colorrangeAVGVFcnn{34.73} & \colorrangeAVGMGcnn{59.48} & \colorrangeAVGMAEcnn{57.18} & \colorrangeAVGSDcnn{59.40} & \colorrangeAVGMeancnn{55.25} \\
 & Warm-Up & \colorrangeAVGVoCocnn{75.47} & \colorrangeAVGSwinUNETRcnn{74.44} & \colorrangeAVGSimCLRcnn{76.30} & \colorrangeAVGVFcnn{76.60} & \colorrangeAVGMGcnn{77.36} & \colorrangeAVGMAEcnn{77.75} & \colorrangeAVGSDcnn{76.40} & \colorrangeAVGMeancnn{76.33} \\
 & Valley & \colorrangeAVGVoCocnn{74.92} & \colorrangeAVGSwinUNETRcnn{74.62} & \colorrangeAVGSimCLRcnn{75.57} & \colorrangeAVGVFcnn{77.84} & \colorrangeAVGMGcnn{76.62} & \colorrangeAVGMAEcnn{76.88} & \colorrangeAVGSDcnn{76.62} & \colorrangeAVGMeancnn{76.15} \\
 & Sawtooth & \colorrangeAVGVoCocnn{75.96} & \colorrangeAVGSwinUNETRcnn{75.31} & \colorrangeAVGSimCLRcnn{76.23} & \colorrangeAVGVFcnn{77.68} & \colorrangeAVGMGcnn{76.66} & \colorrangeAVGMAEcnn{77.50} & \colorrangeAVGSDcnn{76.87} & \colorrangeAVGMeancnn{76.60} \\
\bottomrule
\end{tabular}

    }
\end{table}

\subsection{SSL method performance}
\label{sec:ssl_method_performance}
\paragraph{Segmentation performance}
We compare the fine-tuning results (fine-tuned for 150 epochs) of all pre-training methods for the ResEnc-L CNN architecture and the Primus-M Transformer architecture in \cref{tab:segmentation_results}.
For both architectures, we observe that self-supervised pre-training is able to exceed both, a from-scratch nnU-Net default baseline trained for 1000 epochs and their respective from-scratch trained architecture trained for 150 epochs.
In particular, reconstruction based methods exhibit the best performance, with default MAEs leading the pack. The MAE pre-trained CNN model even exceeds its strong from scratch baseline trained for 1000 epochs in only 150 epochs, and performs even better when fine-tuned for 1000 epochs (see \cref{tab:cnn_1k_epochs_results} in \cref{apx:additional_benchmark_results}). Aside from the MAE pre-training paradigm, this is only achieved by the sparse-autoencoder based S3D pre-training scheme with 1000 epochs of fine-tuning.
\\
Comparing the benefits for each architecture, it can be seen that the Transformer-based Primus-M architecture profits substantially more from pre-training than the ResEnc-L CNN counterpart, however, it shows lower from-scratch results to begin with. Comparing the MAE pre-trained Primus-M to its 1000 epochs from scratch counterpart, it is able to improve about 3.5\% DSC points and almost exceeds the ResEnc-L 1000 epochs from scratch baseline in 150 epochs. However, on singular datasets the MAE pre-trained Primus-M transformer already achieves the highest performance overall (ATL, COS and ACD), including the best pre-trained ResEnc-L. Given the substantial performance discrepancy observed in \citet{wald2025primus} between the two architectures, this provides the first sign of vision transformers reaching state-of-the-art performance in 3D medical image segmentation.\\
However, not all pre-training methods are able to learn useful representations for segmentation. VoCo, SwinUNETR pre-training, and SimCLR (for transformers) do not, or just barely exceed, a from-scratch baseline trained for 150 epochs and can even lead to substantial performance reductions for the transformer architecture likely due to their global pre-training task not learning the local information needed for segmentation.
Moreover, individual datasets can be negatively impacted by any pre-training, as for the ISL and HAN dataset when using a ResEnc-L architecture. This is particularly surprising as Primus-M shows strong improvements with its MAE pre-training on both of these datasets. When filtering low quality images this seems to be partially alleviated (see \cref{tab:datacentric_cnns}), indicating that CNNs may learn different patterns in the presence of noise, which a transformer may be more robust to.\\
Lastly, comparing the longer 1000 epoch fine-tuning schedule (\cref{tab:cnn_1k_epochs_results}) and the shorter 150 epoch fine-tuning schedule of the ResEnc-L, it can be observed that the longer fine-tuning schedule can lead to performance degradation on datasets where pre-training showed strong improvements (SBM, HNT). Simultaneously, it reduces the performance gap to the from-scratch 1000 epoch baseline on cases where pre-training was not as helpful (ISL, HAN, AMO), likely by allowing to forget learned representations and adapt more strongly to the downstream task.\\

\noindent \textbf{Classification performance}
We evaluate the fine-tuned performance of our pre-trained methods for classification -- a more global downstream task -- in \cref{tab:classification_results}. Contrastive pre-training methods like VoCo, SwinUNETR, and SimCLR, which were ineffective for segmentation, achieve the highest classification performance. In contrast, MAE pre-training, which was the strongest for segmentation, performs the worst across both architectures and nearly all datasets. Furthermore, segmentation focused methods like S3D and MG do not transfer well to classification, though they do not fall to the lowest ranks.
These findings reinforce the distinction between global and local feature learning: contrastive methods specialize in global similarity learning, making them well-suited for classification, whereas reconstruction-based objectives prioritize pixel-wise representations, benefiting segmentation. Interestingly, SwinUNETR pre-training, which incorporated both global and local objectives, still converged toward representations favoring classification over segmentation, suggesting a bias in learned feature hierarchies.\\
Additionally, it is to note that from-scratch classification performance is subpar, with ABI barely exceeding random-chance (balanced accuracy of ~50\%). This may be due to the classification pipeline lacking the refinements of the widely adopted nnU-Net framework, due to the lower prevalence of global reasoning tasks in 3D medical imaging.
Thus, while our classification results provide valuable insights, they should be interpreted with caution, particularly compared to our segmentation findings which are more reliable due to the robust benchmarking framework.

\subsection{Fine-tuning methodology}
The ablation results of different CNN fine-tuning strategies in \cref{tab:finetuning_cnns} indicate that the \textit{Sawtooth} fine-tuning strategy consistently achieves the best overall performance, followed closely by the \textit{Warm-Up} scheme. This shows that adapting the encoder, after tuning the newly initialized decoder is crucial to maximize performance. The staged learning rate increases in these schedules likely help to prevent instability and allow for a smoother transition from pre-training to task-specific adaptation. For the out-of-distribution CT datasets, AMO and KIT, the \textit{Default} fine-tuning schedule performs very competitively. This indicates that models pretrained on MRI data require greater modifications when adapted to CT images. The high initial lr allows the models to shift their learned features more effectively to the new domain. \\
Keeping the encoder frozen during fine-tuning significantly reduces performance across all datasets and underscores the importance of encoder adaptation. However, SimCLR is less affected by this constraint. Notably, both VoCo and SimCLR perform almost comparably to other fine-tuning schedules when the encoder is frozen on ATL, w.r.t. the drop on other datasets. This dataset is closest to the pre-training data in the development datasets, as it consists of a single MRI sequence, whereas the other datasets either contain CT scans or multiple MRI sequences as input.
\\
For the Transformer, fine-tuning results in \cref{tab:fine-tuning_transformer} reveal that the \textit{Warm-Up} scheme achieves the best performance across all methods, emphasizing the importance of architecture-specific adaptation. The \textit{Frozen} schedule leads to an even greater performance drop than for CNNs, as the ResEnc-L decoder, despite being lightweight, has significantly more parameters than the "decoder" of Primus-M. In this setting, fine-tuning can be interpreted almost as linear probing, with only back-to-back transposed conv layers (including normalization and activations) being trained, severely limiting decoder capacity. With a frozen encoder, it can be observed that adapting to the ATL dataset reaches the closest performance to other fine-tuning schedules.

\subsection{Data-centric results}
\label{sec:datacentric_results}
We compare the effect of data filtering of the pre-training dataset for CNN MAE pre-training and highlight the measured downstream performance in Appendix \cref{tab:datacentric_cnns}.
It can be observed that filtering the lowest quality images through the Image Quality Score (IQS) shows minor positive effects on overall downstream segmentation performance, despite removing substantial amounts of pre-training data. Only when removing about 66\%, keeping only the highest quality images, a degradation in performance is measured.
Moreover, it can be observed that reducing data diversity by removing all modalities but T1w, T2w and FLAIR images reduces model performance, despite having more pre-training images than the intermediate IQS filtering method which slightly increased performance.
Overall, this indicates that the OpenMind dataset does contain redundant data which may be leveraged by more complex data-centric methods to increase performance further. \\
\noindent\textbf{Effect of anonymization on reconstruction} Many images of the OpenMind dataset were defaced or brain-extracted to guarantee patient privacy. To allow taking these artifacts into account when training with a reconstruction objective, we created masks that allow ignoring such regions. To measure the effect of such artifacts, we compare the performance of reconstruction methods that were or were not penalized for the reconstruction in such regions. (See Appendix \cref{tab:anonymization_mask_effects}). It can be observed that ignoring the reconstruction loss of anonymized regions improves the segmentation performance of the MAE and S3D, showing that taking anonymization into account improves representations. 

\section{Discussion, Limitations and Conclusion}
\label{sec:discussion_and_conclusion}
Our study presents the OpenMind dataset, the largest publicly available 3D medical imaging dataset, alongside a standardized self-supervised learning (SSL) benchmark for 3D medical image analysis. Through extensive experimentation across diverse datasets and across many pre-training paradigms, we demonstrate the superiority of self-supervised pre-training over from-scratch training for 3D medical image analysis, when pre-training on a large-scale dataset.
We show that contrastive methods excel in global tasks such as classification, and reconstruction-based approaches perform well in dense tasks such as segmentation. However, we find that no pre-training method of the 3D vision domain generalizes and performs well in both tasks, emphasizing the need for further SSL method development.\\
\noindent Moreover, we compared state-of-the-art CNNs and Transformers in our benchmark and show for the very first time that a pre-trained Primus-M Transformer can exceed the strong ResEnc-L CNN from-scratch baseline on many datasets and can exceed the strongest pre-trained \mbox{ResEnc-L} CNN on individual datasets. This marks an important milestone for the domain, as transformers from-scratch so far never exceeded a \mbox{ResEnc-L} model~\citep{wald2025primus,isensee2024nnu,bassi2025touchstone}. On average, though, we show that a pre-trained \mbox{ResEnc-L} remains the strongest architecture for now. \\
\noindent Moreover, our ablation results suggest that fine-tuning strategies are a key determinant of the success of pre-trained methods, with bad strategies losing significant performance. This emphasizes the importance of careful adaptation beyond pre-training. Experiments on fine-tuning for longer showed partial degradation of overall performance, indicating the problem of overfitting to downstream tasks. Together, these insights open up research opportunities for developing and adapting Parameter-Efficient Fine-Tuning (PEFT) methods to 3D medical image analysis to address these issues.
Moreover, our fine-tuning experiments showed that keeping the vision encoders frozen shows sub-par performance for all pre-training methods. This emphasizes the lack of generality of currently learned representations and -- again -- highlights the need for better pre-training methods.\\
\noindent While our study provides a lot it also has various limitations: \begin{enumerate*}[label=\roman*)]
    \item Our classification results are not as reliable as the segmentation results, due to the lower interest in classification for 3D medical imaging, requiring further refinements.
    \item Current evidence of the utility of the OpenMind dataset for data-centric research is rather weak, due to the minimal improvements observed requiring further investigation.
    \item We evaluate pre-training for 1000 epochs due to resource limitations. Pre-training for longer may reveal different trends than shown in this study.
\end{enumerate*}\\

\noindent In summary, by open-sourcing the OpenMind dataset and establishing the OpenMind benchmark, we provide a clear assessment of state-of-the-art SSL in 3D medical imaging and offer a reproducible framework for method development. Through them, we aim to accelerate progress in self-supervised learning for 3D medical images and pave the way for new research directions.
{
    \small
    \bibliographystyle{unsrtnat}
    \bibliography{main}
}
\newpage
\appendix
\onecolumn
\section{Benchmark Results: Key Findings and Insights}
\label{apx:additional_benchmark_results}
\begin{table*}
    \centering
    \caption{\textbf{Effects of fine-tuning longer.} We provide additional results of CNNs when fine-tuned not for 150 epochs but 1000 epochs.}
    \label{tab:cnn_1k_epochs_results}
    \resizebox{\linewidth}{!}{\begin{tabular}{l|ccccccccc|ccc|ccc}
\toprule
PT Method & ATL & SBM & ISL & HNT & HAN & MSF & TPC & YBM & COS & ACD & AMO & KIT & ID Mean & OOD Mean & Mean \\
\midrule
nnU-Net def. 1k & \colorrangeATLcnnonek{58.70} & \colorrangeSBMcnnonek{59.98} & \colorrangeISLcnnonek{78.40} & \colorrangeHNTcnnonek{62.98} & \colorrangeHANcnnonek{53.37} & \colorrangeMSFcnnonek{52.19} & \colorrangeTPCcnnonek{79.50} & \colorrangeYBMcnnonek{58.43} & \colorrangeCOScnnonek{46.19} & \colorrangeACDcnnonek{91.10} & \colorrangeAMOcnnonek{88.00} & \colorrangeKITcnnonek{87.21} & \colorrangeIDMeancnnonek{61.08} & \colorrangeOODMeancnnonek{88.77} & \colorrangeMeancnnonek{68.00} \\\midrule
\multicolumn{16}{c}{ResEnc-L (CNN)}\\\midrule
Scratch 1k & \colorrangeATLcnnonek{58.21} & \colorrangeSBMcnnonek{53.43} & \colorrangeISLcnnonek{79.14} & \colorrangeHNTcnnonek{65.75} & \colorrangeHANcnnonek{58.24} & \colorrangeMSFcnnonek{54.90} & \colorrangeTPCcnnonek{79.94} & \colorrangeYBMcnnonek{56.12} & \colorrangeCOScnnonek{71.57} & \colorrangeACDcnnonek{92.09} & \colorrangeAMOcnnonek{88.73} & \colorrangeKITcnnonek{87.48} & \colorrangeIDMeancnnonek{64.15} & \colorrangeOODMeancnnonek{89.43} & \colorrangeMeancnnonek{70.47} \\
VoCo & \colorrangeATLcnnonek{57.52} & \colorrangeSBMcnnonek{58.80} & \colorrangeISLcnnonek{78.08} & \colorrangeHNTcnnonek{61.50} & \colorrangeHANcnnonek{57.43} & \colorrangeMSFcnnonek{54.76} & \colorrangeTPCcnnonek{76.81} & \colorrangeYBMcnnonek{58.05} & \colorrangeCOScnnonek{61.49} & \colorrangeACDcnnonek{91.74} & \colorrangeAMOcnnonek{88.21} & \colorrangeKITcnnonek{87.89} & \colorrangeIDMeancnnonek{62.72} & \colorrangeOODMeancnnonek{89.28} & \colorrangeMeancnnonek{69.36} \\
SwinUNETR & \colorrangeATLcnnonek{56.64} & \colorrangeSBMcnnonek{56.80} & \colorrangeISLcnnonek{77.24} & \colorrangeHNTcnnonek{60.83} & \colorrangeHANcnnonek{55.63} & \colorrangeMSFcnnonek{53.87} & \colorrangeTPCcnnonek{76.17} & \colorrangeYBMcnnonek{57.73} & \colorrangeCOScnnonek{62.16} & \colorrangeACDcnnonek{91.28} & \colorrangeAMOcnnonek{88.22} & \colorrangeKITcnnonek{87.82} & \colorrangeIDMeancnnonek{61.90} & \colorrangeOODMeancnnonek{89.11} & \colorrangeMeancnnonek{68.70} \\
SimCLR & \colorrangeATLcnnonek{57.79} & \colorrangeSBMcnnonek{58.85} & \colorrangeISLcnnonek{78.81} & \colorrangeHNTcnnonek{63.81} & \colorrangeHANcnnonek{58.13} & \colorrangeMSFcnnonek{55.08} & \colorrangeTPCcnnonek{78.75} & \colorrangeYBMcnnonek{59.75} & \colorrangeCOScnnonek{62.45} & \colorrangeACDcnnonek{92.11} & \colorrangeAMOcnnonek{88.58} & \colorrangeKITcnnonek{88.24} & \colorrangeIDMeancnnonek{63.71} & \colorrangeOODMeancnnonek{89.64} & \colorrangeMeancnnonek{70.20} \\
VF & \colorrangeATLcnnonek{59.48} & \colorrangeSBMcnnonek{60.57} & \colorrangeISLcnnonek{78.65} & \colorrangeHNTcnnonek{62.00} & \colorrangeHANcnnonek{55.97} & \colorrangeMSFcnnonek{54.62} & \colorrangeTPCcnnonek{77.76} & \colorrangeYBMcnnonek{59.28} & \colorrangeCOScnnonek{68.29} & \colorrangeACDcnnonek{91.87} & \colorrangeAMOcnnonek{88.59} & \colorrangeKITcnnonek{88.20} & \colorrangeIDMeancnnonek{64.07} & \colorrangeOODMeancnnonek{89.55} & \colorrangeMeancnnonek{70.44} \\
MG & \colorrangeATLcnnonek{57.65} & \colorrangeSBMcnnonek{57.85} & \colorrangeISLcnnonek{77.44} & \colorrangeHNTcnnonek{62.00} & \colorrangeHANcnnonek{57.73} & \colorrangeMSFcnnonek{53.97} & \colorrangeTPCcnnonek{77.79} & \colorrangeYBMcnnonek{58.18} & \colorrangeCOScnnonek{61.27} & \colorrangeACDcnnonek{91.63} & \colorrangeAMOcnnonek{88.59} & \colorrangeKITcnnonek{87.86} & \colorrangeIDMeancnnonek{62.66} & \colorrangeOODMeancnnonek{89.36} & \colorrangeMeancnnonek{69.33} \\
MAE & \colorrangeATLcnnonek{61.04} & \colorrangeSBMcnnonek{62.35} & \colorrangeISLcnnonek{78.71} & \colorrangeHNTcnnonek{64.65} & \colorrangeHANcnnonek{59.48} & \colorrangeMSFcnnonek{54.73} & \colorrangeTPCcnnonek{78.20} & \colorrangeYBMcnnonek{59.47} & \colorrangeCOScnnonek{71.87} & \colorrangeACDcnnonek{92.02} & \colorrangeAMOcnnonek{88.86} & \colorrangeKITcnnonek{88.13} & \colorrangeIDMeancnnonek{65.61} & \colorrangeOODMeancnnonek{89.67} & \colorrangeMeancnnonek{71.63} \\
S3D & \colorrangeATLcnnonek{59.84} & \colorrangeSBMcnnonek{59.84} & \colorrangeISLcnnonek{78.52} & \colorrangeHNTcnnonek{62.76} & \colorrangeHANcnnonek{58.31} & \colorrangeMSFcnnonek{54.37} & \colorrangeTPCcnnonek{78.43} & \colorrangeYBMcnnonek{60.24} & \colorrangeCOScnnonek{72.24} & \colorrangeACDcnnonek{91.93} & \colorrangeAMOcnnonek{88.78} & \colorrangeKITcnnonek{88.10} & \colorrangeIDMeancnnonek{64.95} & \colorrangeOODMeancnnonek{89.61} & \colorrangeMeancnnonek{71.11} \\
\bottomrule
\end{tabular}
}
\end{table*}

\begin{table*}
    \centering
    \caption{\textbf{Fewer data of higher quality provides similar performance.} Removing pre-training data to conduct MAE pre-training on the filtered datasets learns slightly more powerful representations than when using the full dataset. Moreover, reducing diversity of the training data by only training on T1w, T2w and FLAIR images, shows inferior performance.}
    \label{tab:datacentric_cnns}
    \resizebox{\linewidth}{!}{    \begin{tabular}{lll|lllllllll|lll|lll}
\toprule
& & &\multicolumn{15}{c}{Dice Similarity Coefficient (DSC) [\%] on ...}\\
\cline{4-18}
& & &\multicolumn{9}{c|}{Dataset of same anatomical region (ID)} & \multicolumn{3}{c|}{Dataset of OOD region} & \multicolumn{3}{c}{Average across ...} \\ 
PT Method & Data filter & Samples [N] & ATL & SBM & ISL & HNT & HAN & MSF & TPC & YBM & COS & ACD & AMO & KIT & ID  & OOD  & All \\
\midrule
\multirow{5}{*}{MAE} & All & 113.921& \colorrangeATLcnndata{58.25} & \colorrangeSBMcnndata{62.41} & \colorrangeISLcnndata{77.89} & \colorrangeHNTcnndata{66.58} & \colorrangeHANcnndata{55.14} & \colorrangeMSFcnndata{56.84} & \colorrangeTPCcnndata{77.96} & \colorrangeYBMcnndata{60.07} & \colorrangeCOScnndata{70.85} & \colorrangeACDcnndata{91.98} & \colorrangeAMOcnndata{86.78} & \colorrangeKITcnndata{86.12} & \colorrangeIDMeancnndata{65.11} & \colorrangeOODMeancnndata{88.30} & \colorrangeMeancnndata{70.91} \\
 & IQS 3 & 91.952 & \colorrangeATLcnndata{59.33} & \colorrangeSBMcnndata{63.86} & \colorrangeISLcnndata{77.96} & \colorrangeHNTcnndata{66.64} & \colorrangeHANcnndata{52.89} & \colorrangeMSFcnndata{56.52} & \colorrangeTPCcnndata{77.17} & \colorrangeYBMcnndata{59.88} & \colorrangeCOScnndata{73.37} & \colorrangeACDcnndata{92.11} & \colorrangeAMOcnndata{86.61} & \colorrangeKITcnndata{85.66} & \colorrangeIDMeancnndata{65.29} & \colorrangeOODMeancnndata{88.13} & \colorrangeMeancnndata{71.00} \\
  & IQS 2.5 & 65.048& \colorrangeATLcnndata{58.49} & \colorrangeSBMcnndata{65.21} & \colorrangeISLcnndata{78.57} & \colorrangeHNTcnndata{66.42} & \colorrangeHANcnndata{54.40} & \colorrangeMSFcnndata{56.50} & \colorrangeTPCcnndata{78.98} & \colorrangeYBMcnndata{60.28} & \colorrangeCOScnndata{68.90} & \colorrangeACDcnndata{92.08} & \colorrangeAMOcnndata{86.60} & \colorrangeKITcnndata{86.12} & \colorrangeIDMeancnndata{65.31} & \colorrangeOODMeancnndata{88.27} & \colorrangeMeancnndata{71.05} \\
 & IQS 1.5 & 38.225& \colorrangeATLcnndata{58.71} & \colorrangeSBMcnndata{64.69} & \colorrangeISLcnndata{78.24} & \colorrangeHNTcnndata{66.73} & \colorrangeHANcnndata{51.54} & \colorrangeMSFcnndata{56.24} & \colorrangeTPCcnndata{79.00} & \colorrangeYBMcnndata{59.12} & \colorrangeCOScnndata{69.28} & \colorrangeACDcnndata{92.01} & \colorrangeAMOcnndata{86.60} & \colorrangeKITcnndata{86.20} & \colorrangeIDMeancnndata{64.84} & \colorrangeOODMeancnndata{88.27} & \colorrangeMeancnndata{70.70} \\

 & T1w,T2w,FLAIR & 71.314 & \colorrangeATLcnndata{58.91} & \colorrangeSBMcnndata{64.35} & \colorrangeISLcnndata{77.79} & \colorrangeHNTcnndata{65.80} & \colorrangeHANcnndata{51.06} & \colorrangeMSFcnndata{56.96} & \colorrangeTPCcnndata{77.61} & \colorrangeYBMcnndata{59.74} & \colorrangeCOScnndata{69.02} & \colorrangeACDcnndata{92.08} & \colorrangeAMOcnndata{86.57} & \colorrangeKITcnndata{85.84} & \colorrangeIDMeancnndata{64.58} & \colorrangeOODMeancnndata{88.16} & \colorrangeMeancnndata{70.48} \\
\bottomrule
\end{tabular}
}
\end{table*}

\begin{table*}
    \centering
    \caption{\textbf{Effects of considering or ignoring anonymized regions on reconstruction based pre-training methods.} 
    }
    \label{tab:anonymization_mask_effects}
    \resizebox{\linewidth}{!}{
    \begin{tabular}{ll|rrrrrrrrr|rrr|rrr}
\toprule
& &\multicolumn{15}{c}{Dice Similarity Coefficient (DSC) [\%] on ...}\\
\cline{3-17}
& &\multicolumn{9}{c|}{Dataset of same anatomical region (ID)} & \multicolumn{3}{c|}{Dataset of OOD region} & \multicolumn{3}{c}{Average across ...} \\ 
PT Method & Anon. aware & ATL & SBM & ISL & HNT & HAN & MSF & TPC & YBM & COS & ACD & AMO & KIT & ID  & OOD  & All \\
\midrule
\multirow{2}{*}{MAE}  & No & 58.25 & 62.41 & 77.89 & 66.58 & 55.14 & 56.84 & 77.96 & 60.07 & 70.85 & 91.98 & 86.78 & 86.12 & 65.11 & 88.30 & 70.91 \\
  & \cellcolor{lightgray!30} Yes & \cellcolor{lightgray!30} 58.23 & \cellcolor{lightgray!30} 64.77 & \cellcolor{lightgray!30} 78.02 & \cellcolor{lightgray!30} 66.43 & \cellcolor{lightgray!30} 54.60 & \cellcolor{lightgray!30} 56.04 & \cellcolor{lightgray!30} 79.10 & \cellcolor{lightgray!30} 61.14 & \cellcolor{lightgray!30} 72.07 & \cellcolor{lightgray!30} 92.23 & \cellcolor{lightgray!30} 86.77 & \cellcolor{lightgray!30} 86.04 & \cellcolor{lightgray!30} 65.60 & \cellcolor{lightgray!30} 88.34 & \cellcolor{lightgray!30} 71.29 \\\midrule
\multirow{2}{*}{S3D} & No & 58.76 & 64.09 & 78.05 & 65.74 & 52.81 & 56.08 & 78.81 & 59.18 & 66.66 & 92.01 & 86.16 & 86.01 & 64.46 & 88.06 & 70.36 \\
 & \cellcolor{lightgray!30} Yes & \cellcolor{lightgray!30} 59.01 & \cellcolor{lightgray!30} 62.84 & \cellcolor{lightgray!30} 77.92 & \cellcolor{lightgray!30} 66.07 & \cellcolor{lightgray!30} 53.47 & \cellcolor{lightgray!30} 56.01 & \cellcolor{lightgray!30} 77.59 & \cellcolor{lightgray!30} 58.87 & \cellcolor{lightgray!30} 69.45 & \cellcolor{lightgray!30} 92.04 & \cellcolor{lightgray!30} 86.28 & \cellcolor{lightgray!30} 85.80 & \cellcolor{lightgray!30} 64.58 & \cellcolor{lightgray!30} 88.04 & \cellcolor{lightgray!30} 70.45 \\
\bottomrule
\end{tabular}

    }
\end{table*}

\begin{table}
    \centering
    \caption{\textbf{Primus-M Transformer fine-tuning schedule results.} Compared to the ResEnc-L fine-tuning, it can be observed that a normal \textit{Warm-Up} schedule exceeds the \textit{Sawtooth} fine-tuning schedule which proved best for CNNs. \textit{*}: While no-warmup is the default for CNNs, the Primus-M architecture is trained per default with warm-up.}
    \label{tab:fine-tuning_transformer}
    \resizebox{.8\linewidth}{!}{
    \begin{tabular}{ll|lllllll|l}
\toprule
 & PT Method & VoCo & SwinUNETR & SimCLR & VF & MG & MAE & SimMIM & Mean \\
Dataset & FT Schedule &  &  &  &  &  &  &  &  \\
\midrule
\multirow{5}{*}{SBM} & Default* & \colorrangeSBMVoCoTR{42.03} & \colorrangeSBMSwinUNETRTR{47.27} & \colorrangeSBMSimCLRTR{59.61} & \colorrangeSBMVFTR{64.81} & \colorrangeSBMMGTR{62.95} & \colorrangeSBMMAETR{70.86} & \colorrangeSBMSimMIMTR{66.27} & \colorrangeSBMMeanTR{59.12} \\
 & Frozen & \colorrangeSBMVoCoTR{18.56} & \colorrangeSBMSwinUNETRTR{21.50} & \colorrangeSBMSimCLRTR{25.26} & \colorrangeSBMVFTR{39.78} & \colorrangeSBMMGTR{34.62} & \colorrangeSBMMAETR{46.83} & \colorrangeSBMSimMIMTR{43.74} & \colorrangeSBMMeanTR{32.90} \\
 & Warm-Up & \colorrangeSBMVoCoTR{44.16} & \colorrangeSBMSwinUNETRTR{47.65} & \colorrangeSBMSimCLRTR{59.92} & \colorrangeSBMVFTR{66.03} & \colorrangeSBMMGTR{62.79} & \colorrangeSBMMAETR{71.02} & \colorrangeSBMSimMIMTR{66.16} & \colorrangeSBMMeanTR{59.67} \\
 & Valley & \colorrangeSBMVoCoTR{37.55} & \colorrangeSBMSwinUNETRTR{41.64} & \colorrangeSBMSimCLRTR{54.96} & \colorrangeSBMVFTR{65.74} & \colorrangeSBMMGTR{59.11} & \colorrangeSBMMAETR{67.42} & \colorrangeSBMSimMIMTR{67.67} & \colorrangeSBMMeanTR{56.30} \\
 & Sawtooth & \colorrangeSBMVoCoTR{38.19} & \colorrangeSBMSwinUNETRTR{43.20} & \colorrangeSBMSimCLRTR{54.11} & \colorrangeSBMVFTR{64.12} & \colorrangeSBMMGTR{61.27} & \colorrangeSBMMAETR{66.88} & \colorrangeSBMSimMIMTR{68.10} & \colorrangeSBMMeanTR{56.55} \\\midrule
\multirow{5}{*}{ATL} & Default* & \colorrangeATLVoCoTR{48.23} & \colorrangeATLSwinUNETRTR{48.49} & \colorrangeATLSimCLRTR{56.42} & \colorrangeATLVFTR{63.39} & \colorrangeATLMGTR{58.15} & \colorrangeATLMAETR{61.48} & \colorrangeATLSimMIMTR{59.80} & \colorrangeATLMeanTR{56.57} \\
 & Frozen & \colorrangeATLVoCoTR{25.88} & \colorrangeATLSwinUNETRTR{28.99} & \colorrangeATLSimCLRTR{39.90} & \colorrangeATLVFTR{32.91} & \colorrangeATLMGTR{47.18} & \colorrangeATLMAETR{57.57} & \colorrangeATLSimMIMTR{49.59} & \colorrangeATLMeanTR{40.29} \\
 & Warm-Up & \colorrangeATLVoCoTR{48.53} & \colorrangeATLSwinUNETRTR{47.72} & \colorrangeATLSimCLRTR{56.60} & \colorrangeATLVFTR{63.39} & \colorrangeATLMGTR{59.43} & \colorrangeATLMAETR{61.39} & \colorrangeATLSimMIMTR{60.29} & \colorrangeATLMeanTR{56.76} \\
 & Valley & \colorrangeATLVoCoTR{48.48} & \colorrangeATLSwinUNETRTR{48.87} & \colorrangeATLSimCLRTR{51.97} & \colorrangeATLVFTR{63.66} & \colorrangeATLMGTR{57.81} & \colorrangeATLMAETR{60.95} & \colorrangeATLSimMIMTR{60.74} & \colorrangeATLMeanTR{56.07} \\
 & Sawtooth & \colorrangeATLVoCoTR{48.04} & \colorrangeATLSwinUNETRTR{48.79} & \colorrangeATLSimCLRTR{52.81} & \colorrangeATLVFTR{63.00} & \colorrangeATLMGTR{59.17} & \colorrangeATLMAETR{61.19} & \colorrangeATLSimMIMTR{60.69} & \colorrangeATLMeanTR{56.24} \\\midrule
\multirow{5}{*}{AMO} & Default* & \colorrangeAMOVoCoTR{67.28} & \colorrangeAMOSwinUNETRTR{67.43} & \colorrangeAMOSimCLRTR{75.96} & \colorrangeAMOVFTR{85.73} & \colorrangeAMOMGTR{82.95} & \colorrangeAMOMAETR{87.93} & \colorrangeAMOSimMIMTR{88.00} & \colorrangeAMOMeanTR{79.33} \\
 & Frozen & \colorrangeAMOVoCoTR{10.23} & \colorrangeAMOSwinUNETRTR{12.73} & \colorrangeAMOSimCLRTR{21.52} & \colorrangeAMOVFTR{10.74} & \colorrangeAMOMGTR{29.40} & \colorrangeAMOMAETR{37.67} & \colorrangeAMOSimMIMTR{11.02} & \colorrangeAMOMeanTR{19.04} \\
 & Warm-Up & \colorrangeAMOVoCoTR{67.71} & \colorrangeAMOSwinUNETRTR{67.96} & \colorrangeAMOSimCLRTR{76.43} & \colorrangeAMOVFTR{85.40} & \colorrangeAMOMGTR{83.04} & \colorrangeAMOMAETR{88.25} & \colorrangeAMOSimMIMTR{87.79} & \colorrangeAMOMeanTR{79.51} \\
 & Valley & \colorrangeAMOVoCoTR{66.02} & \colorrangeAMOSwinUNETRTR{66.51} & \colorrangeAMOSimCLRTR{75.88} & \colorrangeAMOVFTR{84.81} & \colorrangeAMOMGTR{82.91} & \colorrangeAMOMAETR{87.80} & \colorrangeAMOSimMIMTR{88.39} & \colorrangeAMOMeanTR{78.90} \\
 & Sawtooth & \colorrangeAMOVoCoTR{66.01} & \colorrangeAMOSwinUNETRTR{66.72} & \colorrangeAMOSimCLRTR{76.28} & \colorrangeAMOVFTR{85.03} & \colorrangeAMOMGTR{83.12} & \colorrangeAMOMAETR{87.84} & \colorrangeAMOSimMIMTR{88.22} & \colorrangeAMOMeanTR{79.03} \\\midrule
\multirow{5}{*}{KIT} & Default* & \colorrangeKITVoCoTR{68.31} & \colorrangeKITSwinUNETRTR{68.00} & \colorrangeKITSimCLRTR{77.47} & \colorrangeKITVFTR{82.61} & \colorrangeKITMGTR{82.00} & \colorrangeKITMAETR{86.34} & \colorrangeKITSimMIMTR{84.41} & \colorrangeKITMeanTR{78.45} \\
 & Frozen & \colorrangeKITVoCoTR{22.02} & \colorrangeKITSwinUNETRTR{24.14} & \colorrangeKITSimCLRTR{33.67} & \colorrangeKITVFTR{23.87} & \colorrangeKITMGTR{33.10} & \colorrangeKITMAETR{40.82} & \colorrangeKITSimMIMTR{27.40} & \colorrangeKITMeanTR{29.29} \\
 & Warm-Up & \colorrangeKITVoCoTR{71.52} & \colorrangeKITSwinUNETRTR{69.08} & \colorrangeKITSimCLRTR{80.58} & \colorrangeKITVFTR{82.14} & \colorrangeKITMGTR{80.00} & \colorrangeKITMAETR{86.45} & \colorrangeKITSimMIMTR{83.78} & \colorrangeKITMeanTR{79.08} \\
 & Valley & \colorrangeKITVoCoTR{66.84} & \colorrangeKITSwinUNETRTR{68.72} & \colorrangeKITSimCLRTR{78.70} & \colorrangeKITVFTR{83.02} & \colorrangeKITMGTR{80.72} & \colorrangeKITMAETR{86.29} & \colorrangeKITSimMIMTR{85.54} & \colorrangeKITMeanTR{78.55} \\
 & Sawtooth & \colorrangeKITVoCoTR{66.33} & \colorrangeKITSwinUNETRTR{69.13} & \colorrangeKITSimCLRTR{78.59} & \colorrangeKITVFTR{80.40} & \colorrangeKITMGTR{80.38} & \colorrangeKITMAETR{85.66} & \colorrangeKITSimMIMTR{84.55} & \colorrangeKITMeanTR{77.86} \\\midrule
\multirow{5}{*}{Average} & Default* & \colorrangeAVGVoCoTR{56.46} & \colorrangeAVGSwinUNETRTR{57.80} & \colorrangeAVGSimCLRTR{67.37} & \colorrangeAVGVFTR{74.14} & \colorrangeAVGMGTR{71.51} & \colorrangeAVGMAETR{76.65} & \colorrangeAVGSimMIMTR{74.62} & \colorrangeAVGMeanTR{68.36} \\
 & Frozen & \colorrangeAVGVoCoTR{19.17} & \colorrangeAVGSwinUNETRTR{21.84} & \colorrangeAVGSimCLRTR{30.09} & \colorrangeAVGVFTR{26.83} & \colorrangeAVGMGTR{36.07} & \colorrangeAVGMAETR{45.72} & \colorrangeAVGSimMIMTR{32.94} & \colorrangeAVGMeanTR{30.38} \\
 & Warm-Up & \colorrangeAVGVoCoTR{57.98} & \colorrangeAVGSwinUNETRTR{58.10} & \colorrangeAVGSimCLRTR{68.38} & \colorrangeAVGVFTR{74.24} & \colorrangeAVGMGTR{71.32} & \colorrangeAVGMAETR{76.78} & \colorrangeAVGSimMIMTR{74.50} & \colorrangeAVGMeanTR{68.76} \\
 & Valley & \colorrangeAVGVoCoTR{54.72} & \colorrangeAVGSwinUNETRTR{56.43} & \colorrangeAVGSimCLRTR{65.38} & \colorrangeAVGVFTR{74.31} & \colorrangeAVGMGTR{70.14} & \colorrangeAVGMAETR{75.61} & \colorrangeAVGSimMIMTR{75.58} & \colorrangeAVGMeanTR{67.45} \\
 & Sawtooth & \colorrangeAVGVoCoTR{54.64} & \colorrangeAVGSwinUNETRTR{56.96} & \colorrangeAVGSimCLRTR{65.45} & \colorrangeAVGVFTR{73.14} & \colorrangeAVGMGTR{70.99} & \colorrangeAVGMAETR{75.39} & \colorrangeAVGSimMIMTR{75.39} & \colorrangeAVGMeanTR{67.42} \\
\bottomrule
\end{tabular}

    }
\end{table}

\paragraph{Reconstruction based methods perform best for Segmentation:} 
Our evaluation of SSL pre-training methods demonstrates that reconstruction-based approaches outperform other paradigms in 3D medical image segmentation both for ResEncL and for Primus-M. In particular, default MAE-based pre-training achieves the strongest results, consistently surpassing both a from-scratch nnU-Net baseline trained for 1000 epochs and the respective from-scratch trained architectures fine-tuned for 150 epochs (see \cref{tab:segmentation_results}). Notably, the MAE-pretrained ResEnc-L models not only outperform their from-scratch counterparts in 150 epochs but continue to improve with longer fine-tuning schedules (see \cref{tab:cnn_1k_epochs_results}).

\paragraph{Pre-training accelerates convergence:}
The most significant performance improvements from pre-training are observed within the first 150 epochs of fine-tuning. For instance, Primus-M improves by 4.8 DSC points and ResEnc-L by 2.47 DSC points within this short training duration. This surpasses the relative improvements compared to the full 1000-epoch training from scratch (see \cref{tab:segmentation_results}). This highlights the efficiency of pre-training in enabling faster convergence.

\paragraph{ResEnc-L remains the best overall model:}
Despite the benefits of pre-training across architectures, the ResEnc-L CNN continues to deliver the highest overall segmentation performance. After 150 epochs of fine-tuning, the MAE pre-trained ResEnc-L achieves a mean Dice score of 70.92, slightly outperforming the Primus-M transformer, which reaches 70.42 (see \cref{tab:segmentation_results}).

\paragraph{Pre-Training narrows the performance gap for Primus-M:} While ResEnc-L remains the stronger architecture, pre-training provides a substantial boost to the transformer-based Primus-M, significantly reducing the performance gap. Primus-M benefits far more from pre-training than its CNN counterpart, highlighting the importance of self-supervised learning for architectures that struggle to learn effective representations from scratch. Compared to 1000 epochs of from-scratch training, MAE pre-training followed by 150 epochs of fine-tuning improves Primus-M by 3.43 DSC points, whereas ResEnc-L sees only a 0.44-point gain (see \cref{tab:segmentation_results}). Given the previously observed performance disparity between CNNs and transformers in 3D medical image segmentation \citep{wald2025primus}, these findings suggest that vision transformers, with appropriate pre-training, are increasingly capable of achieving competitive segmentation performance.

\paragraph{Contrastive pre-training offers limited improvements for full model fine-tuning but is best when keeping a frozen encoder for CNNs:} Contrastive learning-based SSL methods—such as VoCo, SwinUNETR pre-training, and SimCLR—provide little to no benefit for segmentation when fine-tuning the entire model (see \cref{tab:segmentation_results}). These approaches either fail to surpass or only slightly improve upon the 150-epoch from-scratch baseline and, in some cases, they even degrade performance, particularly for the transformer-based Primus-M architecture. However, when fine-tuned with a frozen decoder, SimCLR pre-trained models outperform reconstruction-based methods for CNNs in segmentation (see \cref{tab:finetuning_cnns}). 

\paragraph{A frozen Transformer Encoder is worse than a frozen CNN Encoder:} Keeping the encoder frozen during fine-tuning results in an even greater performance drop in performance for Transformers than for CNNs (see \cref{tab:fine-tuning_transformer} and \cref{tab:finetuning_cnns}). While the ResEnc-L decoder allows for some adaptation, the Primus-M transformer has significantly fewer trainable parameters in its light-weight decoder stage. In this setting, fine-tuning is not far away from being linear probing, since only \textit{TransposedConv-Norm-Act} layers are trained. This severely limits the decoder's model capacity, which makes adapting to new tasks more difficult. This underscores the necessity of full fine-tuning for Transformer architectures or the adaptation of Low-Rank adaptation fine-tuning strategies.

\paragraph{Longer fine-tuning improves OOD datasets:} A comparison between the 150 epoch and 1000 epoch fine-tuning schedules reveals a trade-off between adaptation and preservation of pre-trained representations. Longer fine-tuning proves beneficial for datasets where pre-training initially provided little advantage over 1000 epoch from-scratch training (see \cref{tab:cnn_1k_epochs_results} and \cref{tab:segmentation_results}). This effect is particularly pronounced in out-of-distribution (OOD) datasets, such as KIT and AMO, which differ in modality, as well as datasets with a large number of target classes, like HAN and AMO, where convergence generally takes more time. 

\paragraph{Longer fine-tuning can degrade performance where pre-training was already effective:} In cases where pre-training already yielded strong improvements in shorter fine-tuning schedules, extending fine-tuning can lead to performance degradation. Notably, datasets such as SBM and HNT experience a decline when trained for 1000 epochs. These findings suggest that while extended fine-tuning can enhance generalization for OOD datasets, it may also override beneficial pre-trained representations in cases where pre-training was already highly effective.

\paragraph{Contrastive pre-training methods excel in classification:} For ResEnc-L and Primus-M, contrastive pre-training consistently outperforms reconstruction-based approaches in classification tasks (see \cref{tab:classification_results}). While MAE pre-training still provides an improvement over training from scratch, it performs worse than all other methods. Meanwhile, other reconstruction-based methods like MG and S3D achieve moderate classification performance but still fall short of contrastive approaches. This finding reinforces that contrastive learning produces more generalizable global feature representations, making it particularly advantageous for classification tasks.

\paragraph{Balancing data quality and diversity is crucial for optimal performance:}
While the overall impact of filtering for high-quality images is limited, it does lead to slight performance improvements. Using only the top 57\% of images with better quality results in a modest improvement of 0.15 DSC points (see \cref{tab:datacentric_cnns}). However, applying a stricter filter—retaining 62\% of images by selecting only T1w, T2w, and FLAIR sequences—leads to a performance drop of 0.43 DSC points. This suggests that improving image quality can enhance results, but reducing dataset diversity negatively impacts performance. Notably, our quality filtering approach was relatively simple, leaving room for more refined selection strategies.

\paragraph{Impact of anonymization on reconstruction-based methods:} Anonymization modifies image features, potentially impacting reconstruction-based pre-training methods that depend on pixel-level details. Processes like skull-stripping, defacing, or intensity masking may remove valuable anatomical structures, affecting representation learning. Incorporating masks during training to exclude these artificially altered regions from loss calculations slightly improves downstream performance (see \cref{tab:anonymization_mask_effects}). Unsurprisingly, this benefit is observed only for in-distribution datasets covering the head region, as these datasets are directly affected by anonymization processes that alter facial structures or skull regions. In contrast, for out-of-distribution datasets where such modifications are less relevant, anonymization has no noticeable effect.

\section{OpenMind Dataset Details}
\label{apx:dataset_details}
While we provide broad information on the OpenMind dataset in the main manuscript, we extend this and go into more details in the following sections. First we provide more details on the Image Quality Score in \cref{apx:image_quality_score}, provide details of the DWI Preprocessing in \cref{apx:dwi_preprocessing} and lastly provide a broad overview of the dataset distribution in \cref{apx:dataset:metadata}.

\subsection{Image Quality Score}
\label{apx:image_quality_score}
The Image Quality Score is intended as a measurement of the suitability of data for pre-training to remove low-quality images from the pre-training dataset and hopefully increase downstream method performance. We want to denote that the idea of images being of higher or lower utility for pre-training is largely pre-training method-dependent. For example, images that exhibit large amounts of noise will be less suitable for reconstruction-based pre-training methods (e.g. MAEs), due to the presence of noise being impossible to predict for the model. Contrastive methods (e.g. SimCLR) on the other hand may be positively affected due to learning invariance of their representations to such noise.
However, downstream datasets -- which are application focused -- are often composed of high-quality images with minimal noise to optimize outcomes, such as maximizing the accuracy of CyberKnife radiation therapy.
Under these aspects, the Image Quality Score can be interpreted as a measure that quantifies the similarity of the pre-training dataset to the downstream datasets that represent real-world clinical applications.

\paragraph{Image Quality Score quantification}
To create the Image Quality Score (IQS), we inspected two images of each modality in a dataset for all 800 datasets composing the OpenMind dataset\footnote{The image quality assessment process was started with 5 images manually rated, but was reduced to just two images, due to high consistency in image attributes between the different images.}. The resulting Image Quality Score of the two rated images was extended to all images of the same modality within the same dataset, as individual studies showed to be very consistent in their imaging protocols leading to very similar appearance of all images of the same modality within a dataset.
Each of these two images was evaluated on a score from 1 to 5, with 1 indicating clear and sharp imagery and 5 indicating highly compromised image quality, based on their \textit{noisiness} and \textit{blurriness} induced e.g. by subject motion during image acquisition. Artifacts such as visible imaging equipment, the presence of signal voids (visible as shadows, covering a wide area), Gibbs artifacts or other contortions were categorized under \textit{artifact\_level}. The \textit{artifact\_level} was not included for all DWI-derived images, as the preprocessing steps already involved extensive removal of various artifacts and a brain extraction step, see \cref{apx:dwi_preprocessing} for more details. Furthermore, images that were derived from the same DWI image receive the same Image Quality Score. Images where the field of view (FOV) captured only small regions of the brain or where the anatomy covered only a minor portion of the entire volume were also marked. Lastly, raters were able to tag images showing signs of corruption, such as those where missing data was apparent or entirely erroneous. These scores were determined by two independent raters and then aggregated to create the final Image Quality Score (IQS). Specifically, for each dataset and its modalities, the Image Quality Score (IQS) was calculated as follows:\\
\begin{enumerate*}[label=\roman*)]
    \item \textbf{Aggregation of ratings from all raters}: For each image, values on a linear scale (\textit{noisiness}, \textit{blurriness}, and \textit{artifact\_level}) are averaged, while for categorical values, such as the state of corruption or the FOV, the worst possible value is selected to represent the image.\\
    \item\textbf{Calculation of IQS for each image}: First, the average of all numerical values is calculated. However, if the image is marked as corrupted or any numerical value equals 5, the IQS for that image is immediately set to 5. The IQS is increased by one point (capped at 5) if the image's FOV is suboptimal, as described above.\\
    \item\textbf{Modality-Wise IQS Aggregation per Dataset}: For each dataset and modality, each image is assigned the average IQS of these two randomly selected images.
\end{enumerate*}

\subsection{DWI Preprocessing}
\label{apx:dwi_preprocessing}
\begin{figure}
    \centering
    \includegraphics[width=1\linewidth]{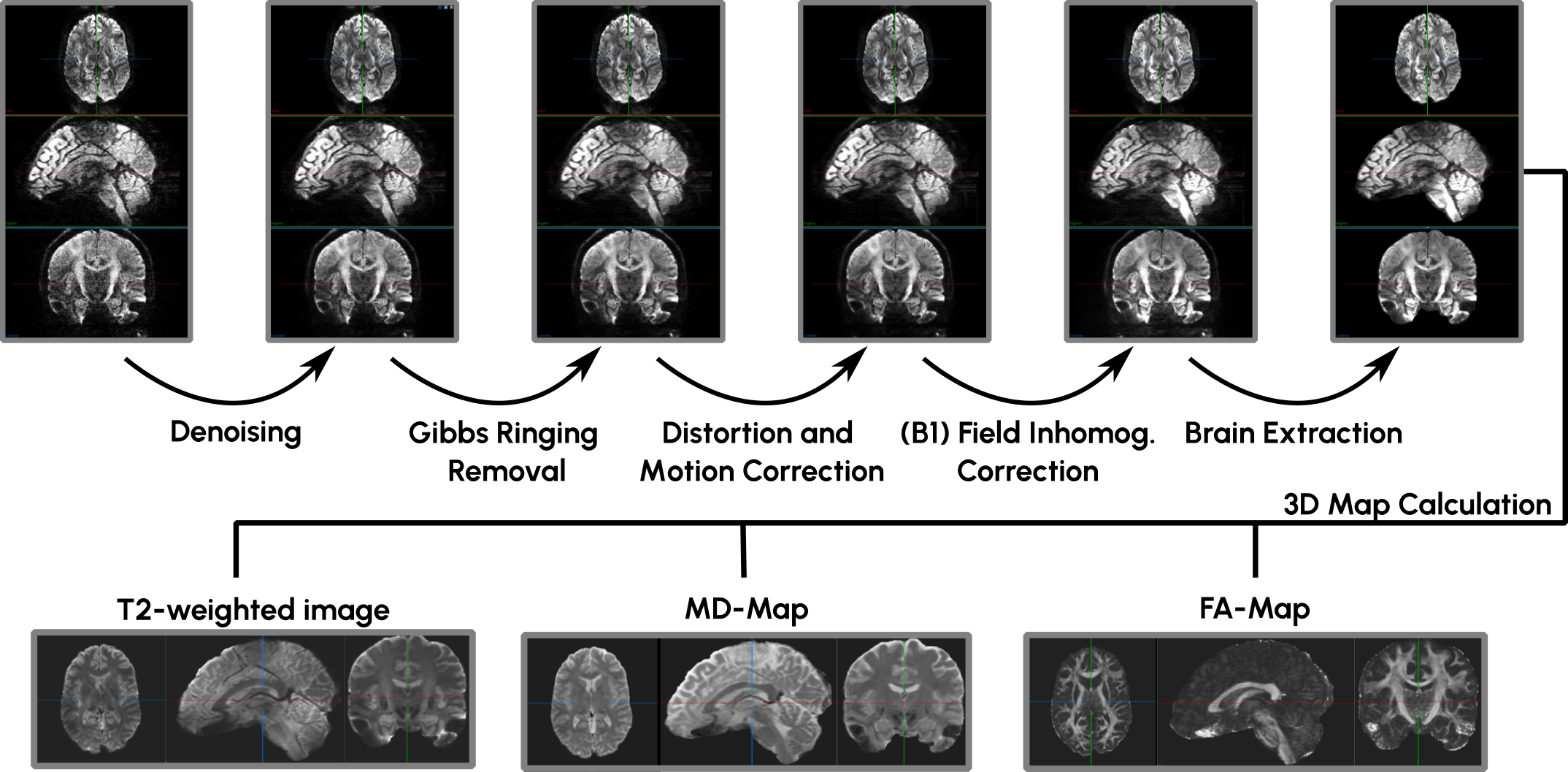}
    \caption{\textbf{DWI preprocessing pipeline:} To derive 3D images from the 4D DWI images, they are processed through six steps, denoising, ringing removal, co-registration, field correction, brain extraction, and lastly 3D derivative creation. Best viewed on a screen to see the differences between steps.}
    \label{fig:dwi_preprocessing_visualization}
\end{figure}
Diffusion-weighted imaging (DWI) measures the diffusion of water molecules in tissue across 3D space. The direction of this diffusion process can be influenced by applying additional magnetic field gradients, allowing researchers to identify tissues that are more or less permeable to water diffusion in specific directions.
Because the diffusion process is directionally dependent, multiple images are acquired using gradients of varying strengths to, for example, quantify anisotropies in diffusion behavior. As a result, DWIs are typically composed of a stack of 3D images, forming complex 4D images that are challenging to process and integrate.
To manage the complexity of these 4D images, they are preprocessed into single 3D image formats that describe specific properties derived from the diffusion measurements in a more interpretable manner. These 3D derivatives of the DWIs allow easier integration into standard SSL pipelines. Specifically, we create T2-weighted, MD, and FA maps from the original 4D DWI stacks.

\noindent The overall preprocessing pipeline is composed of six steps, displayed in \cref{fig:dwi_preprocessing_visualization}:
\begin{enumerate*}[label=\roman*)]
    \item Denoising of each 3D image.
    \item Gibbs ringing artifact removal of each 3D image.
    \item Co-registration of all images to one reference image to correct for potential distortions and motion artifacts.
    \item Field correction of potential B1 field inhomogeneities.
    \item Brain extraction to remove skull and potentially existing face tissue that is not accounted for in the Map creation.
    \item Actual creation of the 3D derivatives, namely a T2-weighted image, an MD-Map and an FA-Map.
\end{enumerate*}

\subsection{Dataset overview}
\label{apx:dataset:metadata}
\begin{figure}
    \centering
    \includegraphics[width=0.7\linewidth]{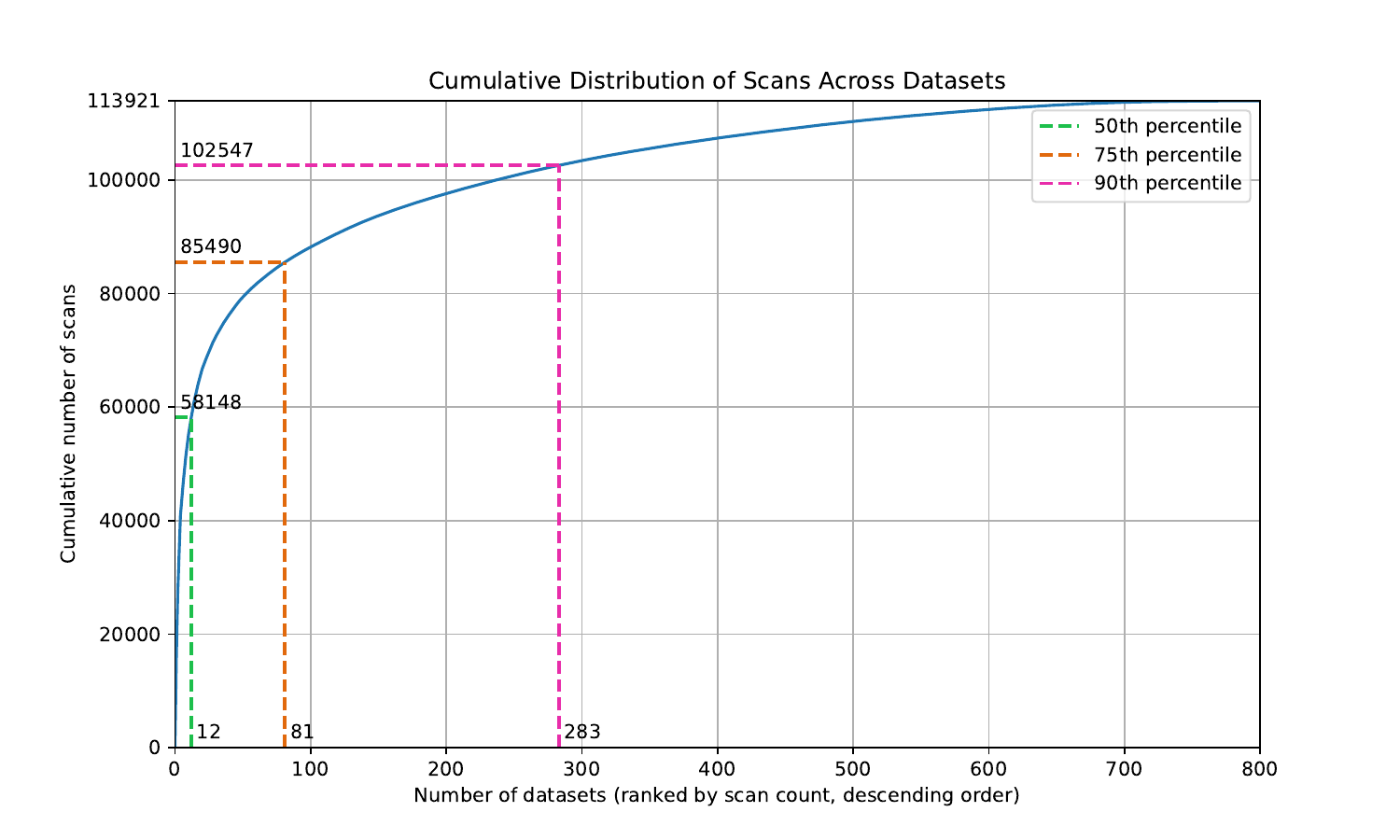}
    \caption{\textbf{Cumulative number of image volumes over number of datasets.} 50\% of all images of the entire OpenMind dataset originate from 12 datasets, while 81 datasets contribute 75\%, 283 contribute 90\% and the remaining 517 contribute the remaining 10\%.}
    \label{fig:cumulative_openmind_dataset_size}
\end{figure}

\paragraph{Dataset origin}
The final OpenMind dataset is comprised of 113,921 3D volumes of 34,191 subjects sourced from exactly 800 datasets. However, the dataset scale differs greatly, with 12 datasets contributing 50\% of all data, 81 contributing 75\% and 283 contributing 90\% of all data, with the remaining 517 contributing the remaining 10\%. While this indicates that many datasets do not contribute a huge quantity of data, they still introduce different niche modalities which may still be of value for the generalization of the pre-trained models. We visualize the cumulative dataset count over an increasing amount of datasets in \cref{fig:cumulative_openmind_dataset_size}.

\paragraph{Image Modalities}
The OpenMind dataset spans 24 Image Modalities -- with all but the PET images being MR image variations. While we report these 24 Image Modalities, we want to emphasize that defining MR modalities or techniques is not trivial -- e.g., T1-weighted MRIs can be acquired with various repetition and echo times, can be acquired by different scanners (changing image appearance drastically) but can still be considered a T1-weighted image. This complicates the assignment of images to specific categories, so the absolute number should be interpreted accordingly. To arrive at our provided number of 24 modalities, we used the provided BIDS modalities (which is not particularly consistent in naming between datasets) and grouped them into the final 24 categories detailed in  \cref{tab:explicit_modality_counts}. Overall, the majority of modalities are the common structural T1w, T2w and FLAIR images, as well as our diffusion derived FA and MD-maps, representing about 86\% of all data, with the remaining images being composed of various other techniques or modalities.
\begin{table}
    \centering
    
    \caption{\textbf{Volume quantity by modality.} Final modality groupings as derived from the OpenNeuro BIDS modality indicators and as known from the own DWI derivatives for FA, MD and T2w maps.}
    \label{tab:explicit_modality_counts}
    \resizebox{.7\linewidth}{!}{\begin{tabular}{llllllll}
\toprule
\rowcolor{lightgray!30}T1w & T2w & FA & MD & FLAIR & MP2RAGE & sbref & ADC \\
42732 & 22999 & 13407 & 13407 & 5583 & 2859 & 1795 & 1757 \\\midrule
\rowcolor{lightgray!30}
TRACE & UNIT1 & inplaneT2 & SWI & UNIT1\_denoised & minIP & PET & T1map \\

1715 & 927 & 892 & 784 & 724 & 687 & 653 & 557 \\\midrule
\rowcolor{lightgray!30}
inplaneT1 & angio & PDw & T2starw & FLASH & T2map & T2starmap & DWI \\
529 & 488 & 473 & 409 & 307 & 106 & 76 & 55 \\
\bottomrule
\end{tabular}}
\end{table}

\begin{figure}
    \centering
    \includegraphics[width=1\linewidth]{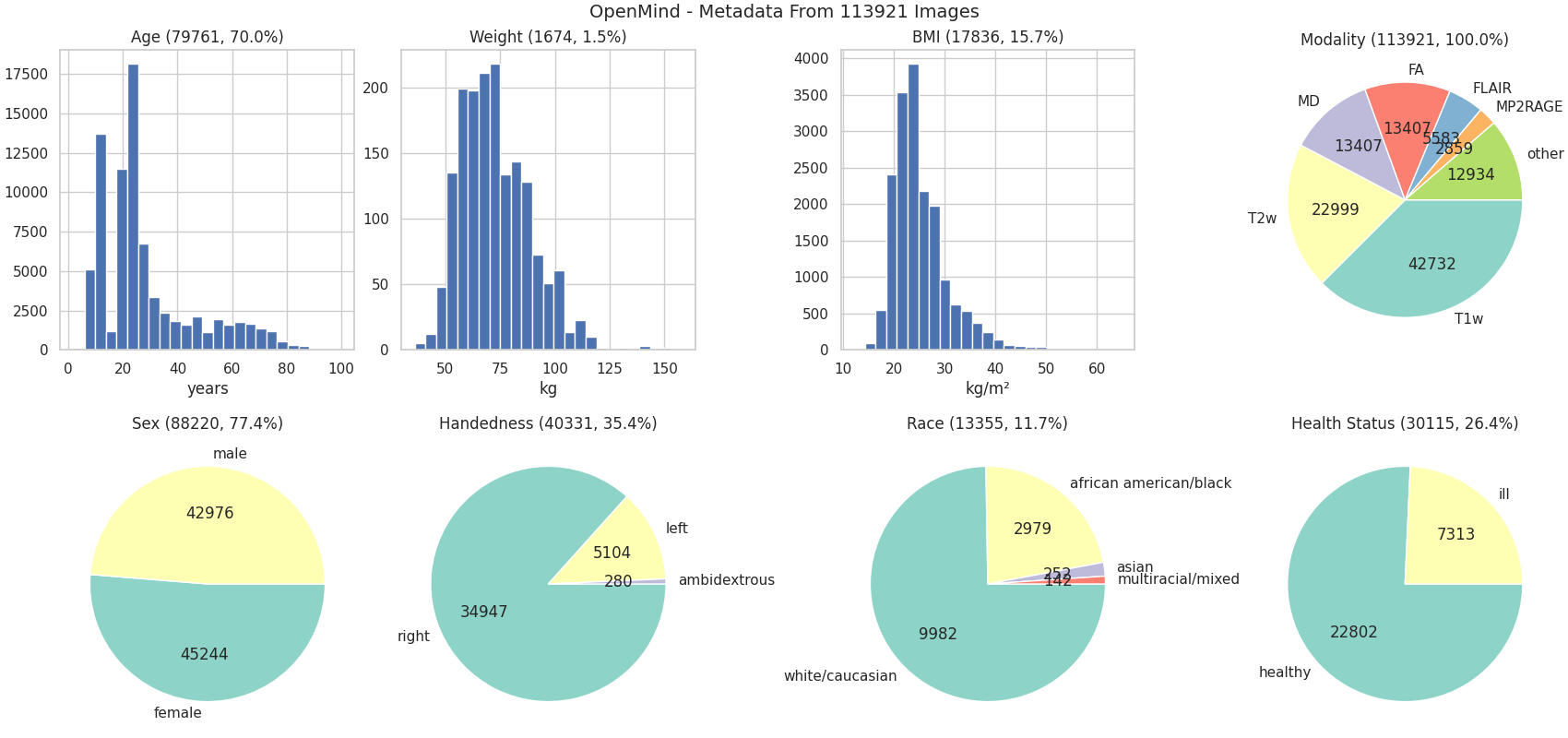}
    \caption{\textbf{Metadata of the OpenMind Dataset:} A total of 113,921 images from 800 datasets were curated and standardized, incorporating key metadata categories such as patient age, weight, BMI, sex, race, and health status, along with imaging modality. To enhance clarity, each pie chart and histogram in the figure only includes scans for which the respective metadata was available. The total number of available cases for each category is displayed above each graphic. Moreover, we denote that this number refers to images and not subjects of which there are only 34k. Therefore, this reflects the image-metadata pair availability instead of a per-patient score, which would not allow knowing the amount of scans with metadata.}
    \label{fig:image_metadata}
\end{figure}

\paragraph{Metadata categories and availability}
Moreover, we display the availability and distribution of metadata information like \textit{age}, \textit{sex}, \textit{bmi}, \textit{handedness}, \textit{health status} and \textit{race} in the OpenMind datasets in \cref{fig:image_metadata}. It can be observed that the majority of datasets provide information about subject sex and age $>70\%$, significantly fewer datasets provided information about the handedness ($35\%$), health status ($26.4\%$), bmi ($15\%$) or race ($11\%$). Even fewer studies recorded subject \textit{weight} ($1.5\%$).
While we only conducted benchmarking of pure self-supervised learning methods that treat each image by itself, there exist pre-training methods that take meta-data into account to provide more efficient pre-training, which could be explored in future work.

\section{Fine-tuning Details}
\label{apx:finetuning_details}
\begin{figure}
    \centering
    \includegraphics[width=1\linewidth]{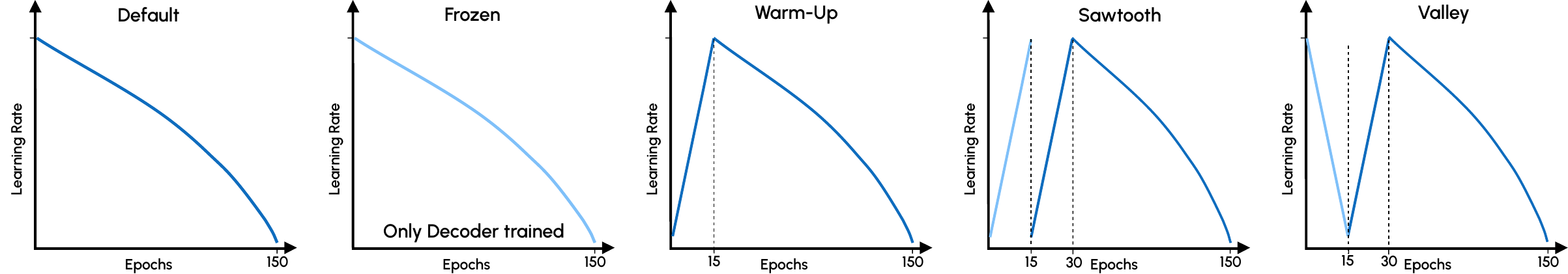}
    \caption{\textbf{Overview of fine-tuning schedules.} The plot illustrates the learning rate (lr) schedules for 150-epoch training runs. For 1000-epoch training, the overall structure remains the same, but each warm-up stage was extended from 15 to 50 epochs. The maximum learning rate was set to 1e-3 for ResEnc-L and 3e-5 for Primus-M. Light blue lines indicate stages where only the decoder was trained, while the encoder remained frozen.}
    \label{fig:finetuning_schedules}
\end{figure}
\subsection{Segmentation Fine-tuning}
All datasets were preprocessed using a fixed cubic 1mm target spacing and z-score normalization mirroring the preprocessing of the pre-training dataset. However, for the pure CT dataset KIT, we applied nnU-Net's default CT normalization, which includes intensity clipping before z-score normalization. Additionally, for TPC, we used nnU-Net's default target spacing, as the 1mm cubic target spacing proved far too coarse for the dataset's thin target structures, which otherwise leads to a significant performance drop. For all nnU-Net default training runs from scratch, we applied the automatically planned, dataset-specific default target spacing and normalization. For fine-tuning, we set the initial learning rate to 1e-3 for ResEnc-L and 3e-5 for Primus-M, reducing the default learning rate of each model by one order of magnitude during the fine-tuning. Below, we describe the different fine-tuning schedules explored in this work:\

\noindent\textbf{Default}: This schedule follows nnU-Net’s standard polynomial decay learning rate (lr) strategy, providing a gradual reduction in learning rate over time.\\
\textbf{Frozen}: The learning rate schedule remains the same as in the Default setting, but training is restricted to the decoder, keeping the encoder parameters fixed.\\
\textbf{Warm-Up}: Fine-tuning begins with a linear increase in the learning rate, allowing for a gradual adaptation before transitioning into the Default schedule.\\
\textbf{Valley}: Training starts by optimizing only the decoder with a linearly decreasing learning rate. This is followed by a warm-up phase, where the entire network is trained using a linearly increasing learning rate, before finally switching to the Default schedule.\\
\textbf{Sawtooth}: This schedule employs a two-stage warm-up. In the first phase, only the decoder is trained with a linearly increasing learning rate while keeping the encoder frozen. In the second phase, the entire network undergoes another warm-up with a linearly increasing learning rate before continuing with the Default schedule.

\noindent For training runs with 150 epochs, each initial warm-up stage lasts 15 epochs, while for 1000-epoch training, each stage is extended to 50 epochs. In all experiments, except for those reported in \cref{tab:finetuning_cnns} and \cref{tab:fine-tuning_transformer}, we use the best-performing fine-tuning schedules: Sawtooth for ResEncL and Warm-Up for Primus-M.
\label{apx:finetuning_details:segmentation}
\subsection{Classification Fine-tuning}
\label{apx:finetuning_details:classification}
All classification fine-tuning was conducted using an adaptation of the \href{https://github.com/MIC-DKFZ/image_classification}{\textit{Image Classification framework}}, which supports 3D classification. Preprocessing in this framework follows the standard nnU-Net preprocessing pipeline, with the important distinction that the entire volumes are resized to a smaller size, as classification is performed on entire images rather than sampled patches to guarantee the presence of potentially important visual cues. \\
\noindent For all datasets, models were trained for 200 dataset epochs (In our classification training, an epoch corresponds to a full dataset epoch.) using a learning rate of 1e-4 and a cosine annealing scheduler with a 20-epoch gradual warm-up applied to both the encoder and classification head. Optimization was performed using AdamW with a weight decay of 1e-2. Given the importance of large batch sizes for classification tasks, gradient accumulation over batches was employed to compensate for memory constraints.
For the MRN dataset, the volumes were resized to $[32\times256\times256]$. The Primus-M architecture was trained with a batch size of 8 and gradient accumulation over 48 batches, while the ResEnc-L architecture used a batch size of 16 with gradient accumulation over 12 batches. On the RSN dataset, volumes were resized to $[160\times192\times192]$. The Primus-M model utilized a batch size of 2 with gradient accumulation over 192 batches, whereas ResEnc-L used a batch size of 4 with accumulation over 48 batches. For the ABI dataset, volumes were resized to $[160\times192\times224]$. Primus-M was trained with a batch size of 2 and gradient accumulation over 48 batches, while ResEnc-L employed a batch size of 4 with accumulation over 96 batches.
Model performance was evaluated using 5-fold cross-validation, and we report Balanced Accuracy and Average Precision as metrics.
Detailed configuration files for each dataset are available within the image classification framework repository.

\section{Self-supervised learning method details}
\label{apx:method_optim}
\begin{figure}
    \centering
    \includegraphics[width=0.9\linewidth]{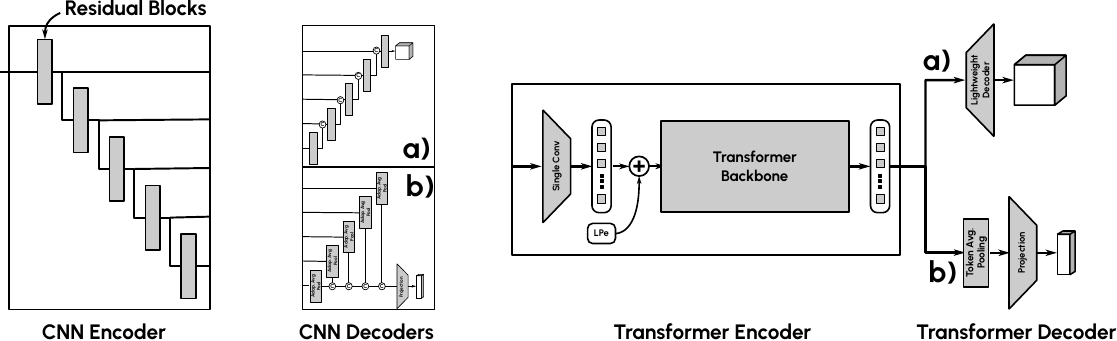}
    \caption{\textbf{Architecture configuration during pre-training.} We provide sketches of the different architectural settings the architectures are pre-trained in, depending on the needs of the respective method, as sometimes a dense and sometimes a global output is required. Since the skip-connections of the ResEnc-L architecture provide outputs on different resolutions, these were averaged through adaptive average pooling and concatenated before projection to the final latent (VoCo, SimCLR). For dense-outputs the highest-resolution output was used (MAE, S3D, VF, MG). The Primus-M architecture does not utilize multi-resolution streams, hence either an iterative up-projection was conducted as proposed in the original paper when requiring dense outputs (MAE, SimMIM, MG, VF), or a token average pooling followed by a linear projection layer was used (VoCo, SimCLR). For the SwinUNETR architecture, which requires dense and global outputs, we follow a) for generating dense outputs, except that no skip connections are used, as described in \cite{tang2022self}, and only a slice/channel for the rotation and contrastive prediction is leveraged, as implemented in their original repository, which we refer to for more details.}
    \label{fig:pretraining_architecture_configurations}
\end{figure}

In this section, we provide a broad overview of the functionality of the different pre-training methods and provide details on hyperparameters of these methods. While some hyperparameters were method specific, we want to denote that we tried to keep hyperparameters as unified as possible to assure fairness. This includes that both ResEnc-L and Primus-M were pre-trained with a batch size of 8 and an input patch size of $[160\times160\times160]$ for the equal amount of 1000 epochs of 250 steps each. Moreover, the ResEnc-L pre-training learning rate was set to 1e-2, weight decay 3e-5, with an SGD optimizer with Nesterov following a PolyLR decay (see "default" in \cref{fig:finetuning_schedules}). The Primus-M architecture was pre-trained with an initial lr pre-training rate of 3e-4, weight decay 5e-2 with an AdamW optimizer following a "Warm-Up" schedule of \cref{fig:finetuning_schedules} with 50 epochs of Warm-Up. 
Moreover, as different pre-training schedules adapt the architecture for the purpose of their method, we provide a visualization of these adaptations of the ResEnc-L UNet architecture and the Primus-M architecture in \cref{fig:pretraining_architecture_configurations}. It is to note that irrespective of the decoder being optimized during the pre-training, which would allow to transfer the pre-trained encoder into the segmentation downstream task, we chose to discard the decoder, as previous work showed this to be slightly superior \citep{wald2024revisiting}. For pre-training, we used a fixed target spacing of cubic 1mm and applied z-score normalization.

\subsection{Volume Contrastive (VoCo)}
\label{apx:method_optim:voco}
\paragraph{Method description}
\citet{wu2024voco} proposed the Volume Contrast (VoCo) framework, which aims to capture contextual relationships in 3D medical images by contrasting different subvolumes within an image. Specifically, given an input volume, they first divide it into $b$ non-overlapping base volumes arranged in a grid-like layout and extract them as individual base crops. In addition, $n$ random crops are extracted from within the same volume. The backbone of the network processes all crops to extract the respective latent features. Following the SimCLR pre-training scheme, the latent features are then passed through a projection head to map them to a different embedding space. 
VoCo optimizes two objectives. The first involves computing similarity logits between a random crop's embedding $p$ and the embeddings of all $n$ base crops $q_i$ using cosine similarity. These predicted similarity scores (ranging from 0 to 1) should align with the actual overlap proportions between the random crop and each base crop.
The second objective introduces a regularization term that encourages the backbone to learn discriminative features between different base crops. This is achieved by calculating the cosine similarity $s_{ij}$ between two different bases $q_i$ and $q_j$ and incorporating it into the loss function, thereby pushing $s_{ij}$ towards a minimum of 0.

\paragraph{Hyperparameter choices}
The final pre-training settings in~\cite{wu2024voco} utilize a grid of $[4\times4\times1]$, yielding $b=16$ base volumes, each with a size of $96^3$, in addition to $n=4$ randomly sampled crops of the same size. This results in a total patch size of $[384\times384\times96]$. Afterwards, all crops are resized to $64^3$. 
With our isometric spacing of $[1\times1\times1]$mm and brain MRIs as our pre-training dataset, a total patch size of $[384\times384\times96]$ proves impractical, given that human brains are far smaller than $38.4cm$ in width. Hence, we skip the downsampling step and directly crop volumes using the final input patch size of the network. Furthermore, we ensure that the grid does not exceed a size of $25.6cm$ in any direction.
As the original hyperparameters proved to be not directly applicable we conducted additional tuning experiments on the validation set of the development datasets, ablating the choice of learning rate (\textit{lr}), weight decay (\textit{wd}), grid number (which indirectly affects the individual total patch size) as well as the number of randomly sampled target crops (\textit{n}), see \cref{tab:voco_optim}. We started the optimization process with $lr=1e^{-2}$, $wd=3e^{-5}$, $n=4$ and a grid of $b=[4\times4\times1]$. We keep the crop size and input patch size at $64^3$ throughout the entire process, leading to an initial total patch size of $[256\times256\times64]$. The final configuration proved to be the same as the initial configuration, with ultimately no changes in the hyperparameter settings. 

\begin{table*}
    \centering
    \caption{\textbf{Optimization of VoCo Hyperparameters.} We optimized Learning Rate, Weight Decay, Grid Size and Number of Random Crops, the most important hyperparameters of the VoCo method in this order. The optimal value for each fine-tuning step is highlighted in gray. \textit{DnC: Did not Converge} during pre-training.
    }
    \label{tab:voco_optim}
    \resizebox{\linewidth}{!}{
    \begin{tabular}{ll|ccc|ccc|ccc|ccc}
\toprule
& & \multicolumn{12}{c}{Dice Similarity Coefficient (DSC) [\%] for different hyperparameters}\\
\cline{3-14}
& &\multicolumn{3}{c|}{1. Learning Rate} &\multicolumn{3}{c|}{2. Weight Decay} &\multicolumn{3}{c|}{3. Grid Size} &\multicolumn{3}{c}{4. Num. Random Crops} \\
\cline{3-14}
 & Ablated value & \cellcolor{lightgray!30}1e-2 & 1e-3 & 1e-4 & 3e-4 & \cellcolor{lightgray!30}3e-5 & 3e-6 & 3x3x1 & \cellcolor{lightgray!30}4x4x1 & 4x4x2 & 2 & \cellcolor{lightgray!30}4 & 8 \\
\cline{3-14}
& Other values &\multicolumn{3}{c|}{wd=3e-5; gr=4x4x1; n=4} &\multicolumn{3}{c|}{lr=1e-2; gr=4x4x1; n=4} &\multicolumn{3}{c|}{lr=1e-2; wd=3e-5; n=4} &\multicolumn{3}{c}{lr=1e-2; wd=3e-5; gr=4x4x1} \\
 
Dataset & Epochs &  &  &  &  &  &  &  &  &  &  &  &  \\
\midrule
\multirow[t]{2}{*}{SBM} & 150 & 71.52 & 70.85 & 71.35 & \textit{DnC}& 71.52 & 65.90 & 69.50 & 71.52 & 72.57 & 69.71 & 71.52 & 72.30 \\
 & \cellcolor{lightgray!30}1000 & \cellcolor{lightgray!30}70.20 & \cellcolor{lightgray!30}73.33 & \cellcolor{lightgray!30}72.98 & \cellcolor{lightgray!30}\textit{DnC}& \cellcolor{lightgray!30}70.20 & \cellcolor{lightgray!30}67.17 & \cellcolor{lightgray!30}72.00 & \cellcolor{lightgray!30}70.20 & \cellcolor{lightgray!30}70.69 & \cellcolor{lightgray!30}71.45 & \cellcolor{lightgray!30}70.20 & \cellcolor{lightgray!30}71.34 \\
\cline{1-14}
\multirow[t]{2}{*}{ATL} & 150 & 60.42 & 59.31 & 60.08 & \textit{DnC}& 60.42 & 56.74 & 60.76 & 60.42 & 58.11 & 59.62 & 60.42 & 60.35 \\
 & \cellcolor{lightgray!30}1000 & \cellcolor{lightgray!30}59.62 & \cellcolor{lightgray!30}59.59 & \cellcolor{lightgray!30}58.32 & \cellcolor{lightgray!30}\textit{DnC}& \cellcolor{lightgray!30}59.62 & \cellcolor{lightgray!30}57.61 & \cellcolor{lightgray!30}58.90 & \cellcolor{lightgray!30}59.62 & \cellcolor{lightgray!30}58.70 & \cellcolor{lightgray!30}58.23 & \cellcolor{lightgray!30}59.62 & \cellcolor{lightgray!30}59.26 \\
\cline{1-14}
\multirow[t]{2}{*}{AMO} & 150 & 86.78 & 85.52 & 82.88 & \textit{DnC}& 86.78 & 84.16 & 86.04 & 86.78 & 86.27 & 86.00 & 86.78 & 86.69 \\
 & \cellcolor{lightgray!30}1000 & \cellcolor{lightgray!30}88.86 & \cellcolor{lightgray!30}88.65 & \cellcolor{lightgray!30}88.57 & \cellcolor{lightgray!30}\textit{DnC}& \cellcolor{lightgray!30}88.86 & \cellcolor{lightgray!30}88.17 & \cellcolor{lightgray!30}88.82 & \cellcolor{lightgray!30}88.86 & \cellcolor{lightgray!30}89.03 & \cellcolor{lightgray!30}88.92 & \cellcolor{lightgray!30}88.86 & \cellcolor{lightgray!30}88.90 \\
\cline{1-14}
\multirow[t]{2}{*}{KIT} & 150 & 85.12 & 82.01 & 80.14 & \textit{DnC}& 85.12 & 80.33 & 83.93 & 85.12 & 83.44 & 83.93 & 85.12 & 83.14 \\
 & \cellcolor{lightgray!30}1000 & \cellcolor{lightgray!30}86.72 & \cellcolor{lightgray!30}84.76 & \cellcolor{lightgray!30}86.40 & \cellcolor{lightgray!30}\textit{DnC}& \cellcolor{lightgray!30}86.72 & \cellcolor{lightgray!30}84.60 & \cellcolor{lightgray!30}84.93 & \cellcolor{lightgray!30}86.72 & \cellcolor{lightgray!30}84.56 & \cellcolor{lightgray!30}85.79 & \cellcolor{lightgray!30}86.72 & \cellcolor{lightgray!30}85.26 \\
\midrule
\midrule
\multirow[t]{2}{*}{Average DSC [\%]} & 150 & 75.96 & 74.42 & 73.61 & \textit{DnC}& 75.96 & 71.78 & 75.06 & 75.96 & 75.10 & 74.82 & 75.96 & 75.62 \\
 & \cellcolor{lightgray!30}1000 & \cellcolor{lightgray!30}76.35 & \cellcolor{lightgray!30}76.58 & \cellcolor{lightgray!30}76.57 & \cellcolor{lightgray!30}\textit{DnC}& \cellcolor{lightgray!30}76.35 & \cellcolor{lightgray!30}74.39 & \cellcolor{lightgray!30}76.16 & \cellcolor{lightgray!30}76.35 & \cellcolor{lightgray!30}75.75 & \cellcolor{lightgray!30}76.10 & \cellcolor{lightgray!30}76.35 & \cellcolor{lightgray!30}76.19 \\
\cline{1-14}
\multirow[t]{2}{*}{Average Rank} & 150 & 1.00 & 2.50 & 2.50 & \textit{DnC}& 1.00 & 2.00 & 2.25 & 1.50 & 2.25 & 2.75 & 1.25 & 2.00 \\
 & \cellcolor{lightgray!30}1000 & \cellcolor{lightgray!30}1.50 & \cellcolor{lightgray!30}2.00 & \cellcolor{lightgray!30}2.50 & \cellcolor{lightgray!30}\textit{DnC}& \cellcolor{lightgray!30}1.00 & \cellcolor{lightgray!30}2.00 & \cellcolor{lightgray!30}2.00 & \cellcolor{lightgray!30}1.75 & \cellcolor{lightgray!30}2.25 & \cellcolor{lightgray!30}1.75 & \cellcolor{lightgray!30}2.00 & \cellcolor{lightgray!30}2.25 \\
\bottomrule
\end{tabular}

    }
\end{table*}

\subsection{Volume Fusion (VF)}
\label{apx:volumefusion}

\paragraph{Method description} Volume Fusion (VF) is a pseudo-segmentation pretext task introduced by \citet{wang2023mis}. Given two separate input volumes $I_f$ and $I_b$ and a voxelwise fusion coefficient map $\alpha$, a fused volume $X$ is generated as follows: $X=\alpha \cdot I_f + (1-\alpha) \cdot I_b$. Each voxel's fusion coefficient $\alpha_i$ is drawn from a discrete set $V=\{0.0,1/K,2/K,\,...,(K-1)/K,1.0\}$, where $K$ denotes the number of nonzero fusion coefficients. Each unique value in $V$ is treated as a class, resulting in $C=K+1$ segmentation classes. Since constructing a fused volume requires two input volumes, each mini-batch sample is formed by drawing a pair of images from the pre-training dataset. The corresponding fusion coefficient map is then generated by sequentially selecting patches of varying sizes at random and assigning each patch a fusion coefficient sampled from $V$. 

\paragraph{Hyperparameter choices} We started the optimization process with $lr=1e^{-2}$, $wd=3e^{-5}$. Furthermore, different values of $K$ were investigated, and an initial value of $K=4$ was employed, following \cite{wang2023mis}. We used center cropping as the sampling strategy, as it promotes spatial consistency and ensures that corresponding anatomical regions align, while also avoiding volumes with large empty patches in our fused volume. All experiments were conducted on the validation set of the development datasets. In the final configuration, the learning rate was adjusted to $1e^{-3}$, while the other settings remained unchanged. Results are shown in \cref{tab:vf_optim}.

\begin{table*}
    \centering
    \caption{\textbf{Optimization of Volume Fusion Hyperparameters.} We optimized Learning Rate, Weight Decay and Number of Nonzero Fusion Coefficients ($K$), the most important hyperparameters of the Volume Fusion method in this order. The optimal value for each fine-tuning step is highlighted in gray.
    }
    \label{tab:vf_optim}
    \resizebox{0.8\linewidth}{!}{
    \begin{tabular}{ll|ccc|ccc|ccc}
\toprule
& & \multicolumn{9}{c}{Dice Similarity Coefficient (DSC) [\%] for different hyperparameters}\\
\cline{3-11}
& &\multicolumn{3}{c|}{1. Learning Rate} &\multicolumn{3}{c|}{2. Weight Decay} &\multicolumn{3}{c}{3. $K$} \\
\cline{3-11}
 & Ablated value & 1e-2 & \cellcolor{lightgray!30}1e-3 & 1e-4 & 3e-4 & \cellcolor{lightgray!30}3e-5 & 3e-6 & 2 & \cellcolor{lightgray!30}4 & 8 \\
\cline{3-11}
& Other values &\multicolumn{3}{c|}{wd=3e-5; $K$=4} &\multicolumn{3}{c|}{lr=1e-3; $K$=4} &\multicolumn{3}{c}{lr=1e-3; wd=3e-5} \\
 
Dataset & Epochs &  &  &  &  &  &  &  &  &  \\
\midrule
\multirow[t]{2}{*}{SBM} & 150 & 73.85 & 73.78 & 75.10 & 76.24 & 73.78 & 74.45 & 74.91 & 73.78 & 75.91 \\
 & 1000 & \cellcolor{lightgray!30}72.43 & \cellcolor{lightgray!30}75.10 & \cellcolor{lightgray!30}72.90 & \cellcolor{lightgray!30}74.09 & \cellcolor{lightgray!30}75.10 & \cellcolor{lightgray!30}72.86 & \cellcolor{lightgray!30}73.33 & \cellcolor{lightgray!30}75.10 & \cellcolor{lightgray!30}75.32 \\
\cline{1-11}
\multirow[t]{2}{*}{ATL} & 150 & 64.99 & 65.07 & 61.58 & 63.61 & 65.07 & 63.84 & 63.64 & 65.07 & 63.71 \\
 & 1000 & \cellcolor{lightgray!30}62.10 & \cellcolor{lightgray!30}62.80 & \cellcolor{lightgray!30}60.72 & \cellcolor{lightgray!30}62.55 & \cellcolor{lightgray!30}62.80 & \cellcolor{lightgray!30}62.62 & \cellcolor{lightgray!30}61.82 & \cellcolor{lightgray!30}62.80 & \cellcolor{lightgray!30}62.69 \\
\cline{1-11}
\multirow[t]{2}{*}{AMO} & 150 & 86.67 & 86.46 & 85.20 & 86.79 & 86.46 & 86.08 & 85.81 & 86.46 & 86.34 \\
 & 1000 & \cellcolor{lightgray!30}89.10 & \cellcolor{lightgray!30}88.95 & \cellcolor{lightgray!30}88.70 & \cellcolor{lightgray!30}89.50 & \cellcolor{lightgray!30}88.95 & \cellcolor{lightgray!30}89.03 & \cellcolor{lightgray!30}89.13 & \cellcolor{lightgray!30}88.95 & \cellcolor{lightgray!30}88.97 \\
\cline{1-11}
\multirow[t]{2}{*}{KIT} & 150 & 84.56 & 85.41 & 82.97 & 83.10 & 85.41 & 84.45 & 83.28 & 85.41 & 84.95 \\
 & 1000 & \cellcolor{lightgray!30}86.21 & \cellcolor{lightgray!30}86.15 & \cellcolor{lightgray!30}85.29 & \cellcolor{lightgray!30}85.65 & \cellcolor{lightgray!30}86.15 & \cellcolor{lightgray!30}86.76 & \cellcolor{lightgray!30}85.25 & \cellcolor{lightgray!30}86.15 & \cellcolor{lightgray!30}85.00 \\
\midrule
\midrule
\multirow[t]{2}{*}{Average DSC [\%]} & 150 & 77.52 & 77.68 & 76.21 & 77.43 & 77.68 & 77.21 & 76.91 & 77.68 & 77.73 \\
 & 1000 & \cellcolor{lightgray!30}77.46 & \cellcolor{lightgray!30}78.25 & \cellcolor{lightgray!30}76.90 & \cellcolor{lightgray!30}77.95 & \cellcolor{lightgray!30}78.25 & \cellcolor{lightgray!30}77.82 & \cellcolor{lightgray!30}77.38 & \cellcolor{lightgray!30}78.25 & \cellcolor{lightgray!30}78.00 \\
\cline{1-11}
\multirow[t]{2}{*}{Average Rank} & 150 & 1.75 & 1.75 & 2.50 & 2.00 & 1.75 & 2.25 & 2.75 & 1.50 & 1.75 \\
 & 1000 & \cellcolor{lightgray!30}1.75 & \cellcolor{lightgray!30}1.50 & \cellcolor{lightgray!30}2.75 & \cellcolor{lightgray!30}2.25 & \cellcolor{lightgray!30}1.75 & \cellcolor{lightgray!30}2.00 & \cellcolor{lightgray!30}2.25 & \cellcolor{lightgray!30}1.75 & \cellcolor{lightgray!30}2.00 \\
\bottomrule
\end{tabular}

    }
\end{table*}

\subsection{Masked Autoencoders (MAE), S3D and SimMIM}
\label{apx:mae}
\paragraph{Method description} Masked autoencoders are a pre-training paradigm that mask a variable amount of the input image and learn to reconstruct the image based on the remaining, non-masked context of the image. Commonly, the pre-training objective only optimizes reconstructing the masked regions, in (normalized) pixel-space through an L2 loss.\\
\textbf{MAEs for Transformers:} MAE pre-training was popularized through \citep{he2022masked} in conjunction with vision transformer architectures for the natural imaging domain (first evaluated in 3D through \citet{chen2023masked}), which leverage the sequence modeling paradigm to improve computational efficiency by discarding the masked tokens effectively reducing the sequence length in the transformer. While this allows the encoder to function very effectively, the mask-tokens need to be re-introduced in order to reconstruct a full image, which is generally done by adding an additional transformer decoder stage, which is discarded when transferring to the downstream task.\\
\textbf{SimMIM for Transformers:} While Transformers are capable of discarding the masked tokens, this is not mandatory. In fact not discarding the tokens, but instead replacing them immediately with learnable Mask Tokens -- introduced in 2D by \citet{xie2022simmim2d} and adapted to 3D by \citet{chen2023masked}-- leads to not needing an additional decoder, which can simplify training. This is generally less computationally efficient and a less common approach.\\
\textbf{MAEs for CNNs:} While keeping or discarding the masked regions in the Transformer is a choice, in CNNs it is mandatory to keep them as no efficient sparse-convolution methods exist to date. Subsequently, CNN MAEs suffer from the same inefficiencies that the SimMIM transformer methods does. Similar to the SimMIM case, the masked regions gradually shrink as the receptive field of the CNNs allows information to leak into the zeroed areas. Moreover, due to a large area in the input being artificially set to zeros, the Batchnorm statistics are influenced by the masking, which may lead to a shift in norm statistics when moving the pre-trained network to downstream tasks, making this approach less prevalent. Despite this, it is still a viable option for CNN pre-training. An exemplary use-case of such an approach in 3D can be seen in \citet{munk2024amaes}.\\
\textbf{S3D for CNNs:}
Given the success of MAEs for Transformers, efforts were made to adapt the CNN architecture to resemble the MAE pre-training paradigm more closely by trying to maintain the masked regions for the entirety of the encoder and filling them at the beginning of the decoder with masked tokens. This also helps to overcome the BatchNorm statistic problems, see \citet{tian2023designing}. This methodology was adopted by \citet{tang2024hyspark,wald2024revisiting} for pre-training 3D medical images. 

\paragraph{Hyperparameter choices}

\textbf{S3D for CNNs:}
We used the publicly available repository provided by \citet{wald2024revisiting}, re-using their final configuration due to them training their method on a dataset from a similar body-region. Notable hyperparameters included a pre-training patch size of $[160\times160\times160]$, learning rate of 1e-2, weight decay of 3e-5, batch size of 8\footnote{In the original repository this was set to 6, but as we developed our models on nodes with 4x40GB A100s we increased this to 8 to evenly distribute the images across all GPUs.}, SGD optimizer with Nesterov and momentum 0.99, randomly sampled masking ratio between [60\%-90\%] with a L2 training loss that is only applied where inputs were masked. Due to the methodological restrictions of needing to mask in the bottleneck, the masked regions consisted of non-overlapping blocks of $[16\times16\times16]$ voxels that were projected up from the bottleneck.\\

\noindent \textbf{MAE for CNNs:}
Our MAE implementation originates from the same repository as S3D~\citep{wald2024revisiting} as the authors had an MAE implemented as an intermediate development stage. We included this implementation in our experiments as is -- again due to the similarity of the dataset. Hyperparameters are identical to those of S3D with the exception of the MAE having a static masking ratio of 75\% and a masking block size of $[16\times16\times16]$.

\noindent \textbf{MAEs for Transformer:}
While existing MAE implementations for Transformers exist (\citep{chen2023masked}), they commonly use default 3D ViT's, which have shown to be far from ResEnc-L CNNs as shown in \citet{isensee2024nnu} or \citet{bassi2025touchstone}. Instead, we use the recent Primus-M architecture, which provided an improved Transformer configuration. Consequently, we draw inspiration from their hyperparameters and used learning rate 3e-4, weight decay 5e-2, AdamW Optimizer with 1e-8 eps and betas (0.9, 0.98) and DropPath 0.2. We adopt randomly dropping tokens, which are of size $[8\times8\times8]$ for Primus with a masking ratio of 75\%. Since the MAE scheme requires a decoder, we introduced a decoder of depth 2, with embedding depth (864) and number of heads (12) -- identical to Primus-M. The masked regions were replaced by learnable embeddings of identical embedding dimensions.\\

\noindent \textbf{SimMIM for Transformer:}
Due to the similarity between the MAE and the SimMIM method, we utilized the same hyperparameters but introduced the learnable mask tokens at the very beginning and introduced no additional decoder since this is not needed for SimMIM.

\subsection{SimCLR}
\label{apx:simclr}
\paragraph{Method decription}
SimCLR (Simple Framework for Contrastive Learning of Visual Representations)~\citep{chen2020simple} is a self-supervised learning approach that relies on contrastive learning to pre-train deep neural networks without labeled data. The core idea behind SimCLR is to maximize the similarity between differently augmented views of the same image while minimizing the similarity between views of different images.
The training framework consists of the following steps:
\begin{enumerate}
    \item Data Augmentation: Each input image is transformed using a set of randomized data augmentations, such as random cropping, color distortion, Gaussian blur, and flipping. These augmentations produce two different views (positive pairs) of the same image.
    \item Encoding: The ResEnc-L or Primus-M backbone processes each augmented image, mapping it to a latent representation.
    \item Projection: The latents are passed through a linear projection head that maps the representations to a lower-dimensional space where the contrastive loss is applied.
    \item Contrastive Loss: The loss function used in SimCLR is the normalized temperature-scaled cross-entropy loss (NT-Xent). This loss encourages the positive pairs (two augmentations of the same image) to be close in representation space while pushing away representations of different images (negative pairs).
\end{enumerate}
By minimizing this training objective, the model learns to generate useful feature representations by learning to become invariant to the augmentations, learning to focus on the semantic meaning, while learning that different categories should be different. After training, the encoder is fine-tuned, and the projection head is discarded.\\

\paragraph{Hyperparameter choices}
Contrastive pre-training methods generally require larger batch sizes as they need to learn what to be similar and dissimilar to. We train with a batch size of 32, receiving input crops of $[192\times192\times64]$ of which we create two augmented sub-crops of dimensions $[64\times64\times64]$ per input crop with a minimal crop-overlap of 50\%. Moreover, the NTXent loss uses a temperature scale of 0.5 as well as a cosine similarity function between the latents. To create the augmented versions, we employ \textit{GaussianNoise}, \textit{GaussianBlur}, \textit{BrightnessMultiplicative}, \textit{ContrastAugmentation}, \textit{SimulateLowResolution}, \textit{Gamma}, \textit{Mirror} and \textit{Rotate90Deg} transforms of the common batchgenerators framework\footnote{https://github.com/MIC-DKFZ/batchgenerators}. We refer to the repository for explicit parametrization and potential repetitions of the same augmentations.

\subsection{SwinUNETR}
\label{apx:swinunetr}
\paragraph{Method description}
SwinUNETR was proposed as a pre-training method for the identically named SwinUNETR architecture and is composed of three components: Image inpainting, Rotation prediction as well as a contrastive training objective. The inpainting itself is a simple L1 loss applied for a masked out image region, the rotation is a rotation of 0\textdegree, 90\textdegree, 180\textdegree~or 270\textdegree~degrees along the z-axis with an MLP used to classify the applied rotation. Lastly, the contrastive coding enforces the linearly projected representations of the encoder to be highly similar or dissimilar if the two sub-volumes belong to the same or a different image, respectively. These three losses are combined with an equal weighting to form the SwinUNETR pre-training. 

\paragraph{Hyperparameter choices}
Adhering to the original hyperparameters we calculate the in-painting loss with an L1 loss function, the rotation classification through a cross entropy loss and the contrastive loss and aggregate the final loss through aggregating the losses with weights $\lambda_{rot}=\lambda_{inpaint}=\lambda_{contrast}=1$. For the rotation objective, we follow the description in the paper and rotate along the z-axis. For the contrastive objective, we followed the original implementation, which uses random rotations and random cut-outs with a maximum removal rate of 60\% (In the paper, they mention a drop rate of 30\% but override this in the implementation).
As no other model-specific hyperparameters had to be chosen, we trained the model with a batch size of 8 and an input image shape of $[160\times160\times160]$, which was internally doubled by SwinUNETR through the contrastive loss objective's augmentations, making it train substantially longer than the other methods. 
\alldatasetcite
\end{document}